\documentclass{article}

\usepackage[preprint]{neurips_2026}

\usepackage[utf8]{inputenc} 
\usepackage[T1]{fontenc}    
\usepackage{hyperref}       
\usepackage{url}            
\usepackage{booktabs}       
\usepackage{amsfonts}       
\usepackage{nicefrac}       
\usepackage{microtype}      
\usepackage{xcolor}         

\usepackage{amsmath}
\usepackage{amssymb}
\usepackage{mathtools}
\usepackage{amsthm}
\usepackage{subcaption} 
\usepackage{graphicx}
\usepackage{listings}
\usepackage{xcolor}
\usepackage{booktabs}
\usepackage{array}
\usepackage[table]{xcolor} 
\definecolor{codegreen}{rgb}{0,0.6,0}
\definecolor{codegray}{rgb}{0.5,0.5,0.5}
\definecolor{codepurple}{rgb}{0.58,0,0.82}
\definecolor{backcolour}{rgb}{0.95,0.95,0.92}

\lstdefinestyle{mystyle}{
    backgroundcolor=\color{backcolour},   
    commentstyle=\color{codegreen},
    keywordstyle=\color{magenta},
    numberstyle=\tiny\color{codegray},
    stringstyle=\color{codepurple},
    basicstyle=\ttfamily\small,
    breakatwhitespace=false,         
    breaklines=true,                 
    captionpos=b,                    
    keepspaces=true,                 
    numbers=left,                    
    numbersep=5pt,                  
    showspaces=false,                
    showstringspaces=false,
    showtabs=false,                  
    tabsize=2
}

\usepackage{wrapfig}

\newcommand{\z}{z}

\DeclareMathOperator{\EX}{\mathbb{E}}

\newcommand{\relu}{\texttt{ReLU}}

\theoremstyle{plain}
\newtheorem{theorem}{Theorem}[section]

\theoremstyle{definition}

\theoremstyle{remark}

\usepackage[textsize=tiny]{todonotes}

\usepackage[utf8]{inputenc} 
\usepackage[T1]{fontenc}    
\usepackage{hyperref}       
\usepackage{url}            
\usepackage{booktabs}       
\usepackage{amsfonts}       
\usepackage{nicefrac}       
\usepackage{microtype}      
\usepackage{geometry} 

\usepackage{tikz}
\usetikzlibrary{positioning}

\usepackage{minitoc} 
\doparttoc                      
\mtcsetdepth{parttoc}{2}        
\mtcsettitle{parttoc}{Appendix Contents}  

\usepackage[capitalize,noabbrev]{cleveref}

\title{Hadamard Representation: Scaffolding Performance Across Model-free RL}

\author{%
  Jacob E. Kooi \\
  \texttt{jacobkooi92@gmail.com} \\
  \And
  Zhao Yang \\
  \texttt{z.yang3@vu.nl} \\
  \And
  Mark Hoogendoorn \\
  \texttt{m.hoogendoorn@vu.nl} \\
  \And
  Vincent François-Lavet \\
  \texttt{vincent.francoislavet@vu.nl} \\
  \AND
  \textnormal{Vrije Universiteit Amsterdam}
}

\begin{document}
\faketableofcontents  

\maketitle

\begin{abstract}
Deep reinforcement learning agents progressively lose representational capacity during training: neurons become dormant, removing active capacity from the network, and effective rank collapses, leaving surviving neurons redundant. Existing remedies such as periodic resets, and special neural network architectures, are largely algorithm- or domain-specific. We propose a simple architectural fix, the Hadamard Representation (HR), which replaces a standard hidden layer with the element-wise product of two independently parameterized layers. HR operates through two complementary mechanisms. First, it reduces the probability of a neuron becoming dormant, which is particularly valuable for continuously differentiable activations such as $\tanh$: unlike dormant ReLU neurons, which are effectively pruned, saturated $\tanh$ neurons silently corrupt downstream layers by turning their outgoing weights into fixed biases. Second, independently of dormancy, the multiplicative structure captures richer feature interactions and increases effective rank without widening the layer. We evaluate HR across five algorithms and three domains: DQN, PPO, and PQN on pixel-based discrete-action Atari, SimbaV2 on state-based continuous control, and MR.Q on visual continuous control. HR consistently improves performance over the strong baselines without any hyperparameter tuning, and gains persist against parameter-matched wider variants, ruling out parameter count as an alternative explanation.
\end{abstract}

\section{Introduction}
\label{sec:intro}

\begin{figure}[!htb]
    \centering
    \includegraphics[width=0.9\linewidth]{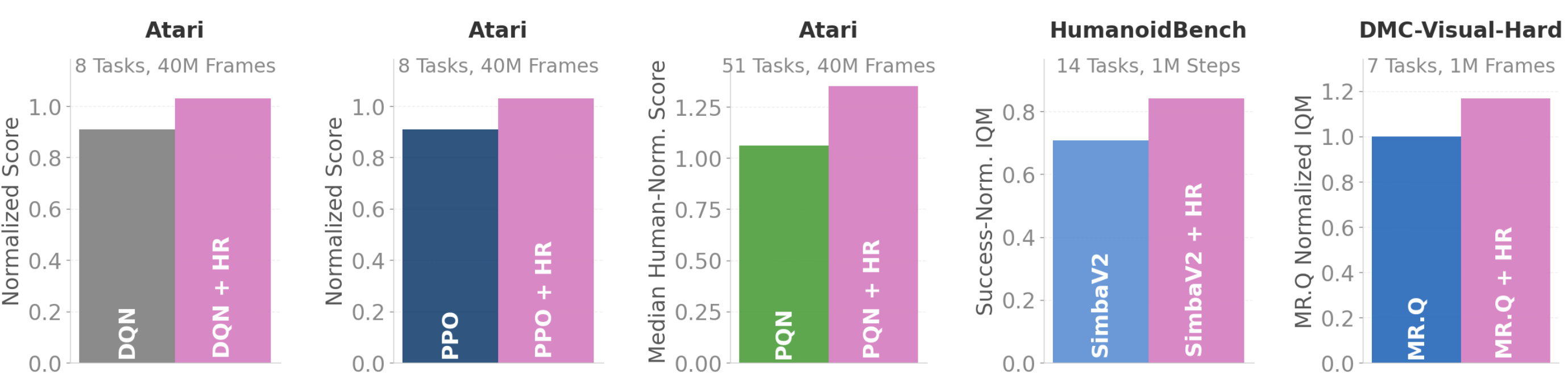}
    \caption{Improvements with Hadamard Representation (HR) over the baselines across five RL algorithms and three domains.}
    \label{fig:first_plot}
\end{figure}

Deep reinforcement learning agents progressively lose representational capacity as training unfolds: neurons become dormant~\citep{PabloSamuel_Dying}, effective rank collapses~\citep{Lyle_Capacity_loss}, and learned features become impoverished~\citep{Dohare2024_nature_continualbackprop}. Existing remedies such as auxiliary losses~\citep{schwarzer2020data}, periodic resets~\citep{PabloSamuel_Dying}, and normalization~\citep{lee2024simba} address these symptoms through algorithm- or domain-specific adjustments. In this work, we instead revisit the problem from an architectural angle, starting with the behavior of dormant neurons under different activation functions.

The dying $\relu$ phenomenon in RL is well-documented~\citep{PabloSamuel_Dying, dying_relu2, Tanh_Rank_Study_TMLR}, but $\tanh$ has received far less attention despite producing representations with fewer dormant neurons and higher effective rank~\citep{Tanh_Rank_Study_TMLR}. If the dormant count were the full story, $\tanh$ should outperform $\relu$ in RL. Empirically, it does not. This suggests that not all dormant neurons are equal: a dormant $\relu$ outputs zero and is effectively pruned, whereas a saturated $\tanh$ neuron collapses to $\pm 1$ and converts its outgoing weights into a fixed, input-independent bias that silently distorts downstream layers.

Motivated by this observation, we study a simple architectural change that reduces the probability of a $\tanh$ neuron becoming dormant in the first place: the \textbf{Hadamard Representation (HR)}, which replaces a standard hidden layer with the element-wise product of two independently parameterized activations,
\begin{equation}
    \mathbf{z}(\mathbf{x}) = f(\mathbf{A}_1 \mathbf{x} + \mathbf{b}_1) \odot f(\mathbf{A}_2 \mathbf{x} + \mathbf{b}_2),
\end{equation}
where $\mathbf{x}$ is the input to the layer, $\mathbf{z}$ its output, $f$ the activation function, and $(\mathbf{A}_i, \mathbf{b}_i)$ two independent sets of weights and biases. Under HR, product saturation requires both branches to saturate simultaneously, which reduces the per-neuron collapse probability for $\tanh$. Beyond saturation resistance, HR offers a second, complementary effect: the product of two independently parameterized activations captures second-order interactions between two views of the input, creating a richer function class at the same layer width. Similar multiplicative structures appear in gating mechanisms such as GLUs~\citep{Gated_Linear_Unit_Dauphin_Swiglu} and highway networks~\citep{Highway}, but in HR both branches are full activation outputs rather than one being a scalar gate, so the structure remains symmetric. Even when dormant neurons are already rare, this multiplicative expressiveness can provide gains.

We evaluate HR across five algorithms and three domains, see~\cref{fig:first_plot}. In pixel-based discrete-action Atari~\citep{bellemare2013arcade} domain, HR delivers over 10\% gains for DQN~\citep{mnih2013playing} and PPO~\citep{schulman2017proximal} across 8 games and shows consistent scaling on 51 games with PQN~\citep{pqn}. In continuous control, HR improves state-of-the-art methods like SimbaV2~\citep{lee2025hyperspherical} on state-based domain HumanoidBench~\citep{sferrazza2024humanoidbench} and MR.Q~\citep{fujimoto2025towards} on visual-based domain DMC-Visual~\citep{Tassa2018DeepMindSuite} \textbf{without} any hyperparameter tuning.

Our contributions are as follows:
\begin{itemize}
    \item We identify a failure mode unique to smooth activations in RL: saturated $\tanh$ neurons silently convert their outgoing weights into hidden biases, a strictly more harmful failure mode than $\relu$ pruning.
    \item We study the Hadamard Representation (HR), a drop-in multiplicative layer that reduces the per-neuron collapse probability for $\tanh$, while amplifying it for ReLU. Independently of its effect on dormancy, HR also increases the effective rank of hidden representations, which benefits architectures where dormant neurons are already rare.
    \item We show consistent gains across three diverse domains: pixel-based Atari with discrete actions (DQN, PPO, PQN), state-based continuous control (SimbaV2 on HumanoidBench), and visual continuous control (MR.Q on DMC-Visual). PQN, SimbaV2, and MR.Q are among the strongest methods in their respective domains, and HR improves them further without any hyperparameter tuning.
\end{itemize}

\section{Related Work}

We organize related work along two axes: (i) representational capacity loss in RL and its proposed mitigation, and (ii) architectural innovations in model-free deep RL.

\paragraph{Capacity loss in RL.}
A growing body of work has documented how deep RL networks progressively lose representational capacity during training. Prior work has investigated the role of sparse representations in continuous control~\citep{liu2019utility} and introduced effective rank as a measure of network expressiveness in RL~\citep{Effective_rank_kumar_2019}. Subsequent analysis showed that $\tanh$ activations preserve high effective rank and resist rank collapse~\citep{Tanh_Rank_Study_TMLR}, and capacity loss has been characterized as a general phenomenon in which effective network capacity decays as training progresses~\citep{Lyle_Capacity_loss}. In pixel-based robotic control, normalization combined with action penalization has been shown to reduce variance and capacity loss~\citep{Bjorck2021IsControl}.

Several interventions have been proposed to mitigate these pathologies. In the sample-efficient regime, periodic network resets counteract the primacy bias~\citep{Network_Resets_RL_Nikishin_Primacy_Bias}, and the dying ReLU phenomenon has been characterized and mitigated through targeted resets in DQN~\citep{PabloSamuel_Dying}. For longer training horizons, plasticity injection~\citep{Plasticity_Injection_Nikishin_Neurips_2023}, rational activations~\citep{rational_activations, delfosse2024adaptive}, and continual backpropagation~\citep{Dohare2024_nature_continualbackprop} have been proposed to preserve plasticity. A complementary line of work shows that a substantial fraction of network capacity is unnecessary for effective training~\citep{Offline_Sparse_RL_Single_shot_pruning_precup, Graesser_Pablo_Sparse_RL, Sokar_Dynamic_Sparse_IJCAI, Sparse_Deep_RL_Tan_Hu_Arxiv, Pruned_network_rl}, offering a plausible explanation for why ReLU networks achieve strong performance despite a significant fraction of dormant neurons. Our approach is complementary to these lines of work: rather than proposing an algorithmic fix that targets symptoms of capacity loss, we introduce a minimal architectural change that directly reduces the probability of neuron dormancy.

\paragraph{Architecture in model-free RL.}
An early empirical study of activation functions and architectures in RL is given by Henderson et al.~\citep{henderson2018deep}. Subsequent work has explored architectural design across the three main model-free settings. \textbf{Pixel-based Atari.} Impala~\citep{espeholt2018impala} introduced deep ResNet~\citep{he2016residual} blocks to enable high data efficiency under distributional training, and BBF~\citep{schwarzer2023bigger} further widened the Impala architecture to achieve strong performance on Atari-100k~\citep{ye2021mastering}. A separate line of work has investigated pooling: Hadamax combines max-pooling with Hadamard representations~\citep{kooi2025hadamax}, while Impoola~\citep{trumpp2025impoola} and related approaches~\citep{sokar2025mind} use global average pooling, both tailored for Atari encoders. \textbf{State-based continuous control.} Purpose-built architectures have demonstrated strong performance through carefully chosen normalization: LayerNorm~\citep{nauman2024bigger, lee2024simba}, RSNorm~\citep{lee2025hyperspherical}, and BatchNorm~\citep{bhatt2019crossq, palenicek2025scaling, palenicek2025xqc}. \textbf{Visual continuous control.} Architectural innovations are comparatively scarce, and most gains have instead come from data augmentation~\citep{kostrikov2020image, yarats2021mastering}, improved representation-learning objectives~\citep{schwarzer2020data, fujimoto2025towards}, or better exploration~\citep{xu2023drm}. HR is orthogonal to these directions and can be combined with them: it is a drop-in layer-level modification that does not alter the encoder topology, normalization scheme, or training objective.

\section{Preliminaries}

\paragraph{Reinforcement learning.}
We consider an agent acting within its environment as a Markov Decision Process (MDP) defined as a tuple $(\mathcal{S}, \mathcal{A}, T, R, \gamma)$~\citep{andrew2018reinforcement,vincent2018introduction}. $\mathcal{S}$ is the state space, $\mathcal{A}$ is the action space, $T: \mathcal{S} \times \mathcal{A} \rightarrow \mathbb{P}(\mathcal{S})$ is the transition function, $R: \mathcal{S} \times \mathcal{A} \rightarrow \mathbb{R}$ is the reward function, and $\gamma \in [0,1)$ is the discount factor. The agent's goal is to learn a policy $\pi: \mathcal{S} \rightarrow \mathcal{A}$ that maximizes the expected discounted return $V^\pi(s) = \EX_{\tau}[\sum_{t=0}^{\infty} \gamma^{t} R(s_{t}, a_{t}) \mid s_0 = s]$, where $\tau$ is a trajectory following $\pi$.

However, in many practical settings, including the pixel-based tasks investigated in this work, the agent lacks direct access to the underlying environment state $s_t$. These scenarios are more accurately modeled as a Partially Observable Markov Decision Process (POMDP)~\citep{kaelbling1998planning}, defined by the augmented tuple $(\mathcal{S}, \mathcal{A}, T, R, \Omega, O, \gamma)$. In this framework, $\Omega$ is the observation space and $O: \mathcal{S} \times \mathcal{A} \rightarrow \mathbb{P}(\Omega)$ is the observation function, characterizing the probability of perceiving $o_t \in \Omega$ given the current state and preceding action. In Atari or visual continuous control that we are used in this work, it is standard practice to stack multiple consecutive frames to form a state proxy, effectively approximating the Markov property by incorporating the temporal context necessary to resolve partial observability.

\paragraph{Dormant neurons.}
A neuron $\alpha_i$ in a hidden layer of dimension $w$ is considered dormant if it produces an approximately constant output across observations, indicating a loss of plasticity. Specifically, a neuron is dormant if $\alpha_i \approx \Omega$ for all $o_t \in B$, where $o_t$ is an observation in the minibatch $B$ and $\Omega$ is the saturation limit of the activation function: $\Omega = 0$ for $\relu$ and $\Omega \in \{-1, +1\}$ for $\tanh$. For $\relu$, this reduces to a direct check of whether activations are identically zero across the batch. For $\tanh$, an approximate equality is necessary since the function reaches $\pm 1$ only asymptotically. We operationalize this via kernel density estimation (KDE)~\citep{silverman1986density} on per-neuron activation distributions, flagging a neuron as dormant when the peak of its estimated density exceeds a threshold. 
Full details on bandwidth selection and KDE computation are provided in Appendix~\ref{app:kde}.

\paragraph{Effective rank.}
Following Kumar et al.~\citep{Effective_rank_kumar_2019}, the effective rank of a feature matrix $\Phi$ at threshold $\delta$ is defined as:
\begin{equation}
    \mathrm{srank}_{\delta}(\Phi) = \min \left\{ k : \frac{\sum_{i=1}^{k} \sigma_i(\Phi)}{\sum_{i=1}^{d} \sigma_i(\Phi)} \geq 1 - \delta \right\},
\end{equation}
where $\{\sigma_i(\Phi)\}$ are the singular values of $\Phi$ sorted in descending order, i.e., $\sigma_1 \geq \dots \geq \sigma_d \geq 0$. 
Intuitively, $\mathrm{srank}_\delta$ represents the minimum number of orthogonal singular-value directions required to capture $(1 - \delta)$ of the total spectral mass. This corresponds to the number of ``effective'' unique components that form the basis for linearly approximating the learned representation; a higher value indicates that a greater number of dimensions carry distinct, useful signals. Full details regarding the computation of effective rank can be found in~\cref{app:rank_calc}.

\section{From Dormant Neurons to Hadamard Representations}
\label{sec:method}

\begin{figure}[!htb]
    \centering
    \includegraphics[width=0.9\linewidth]{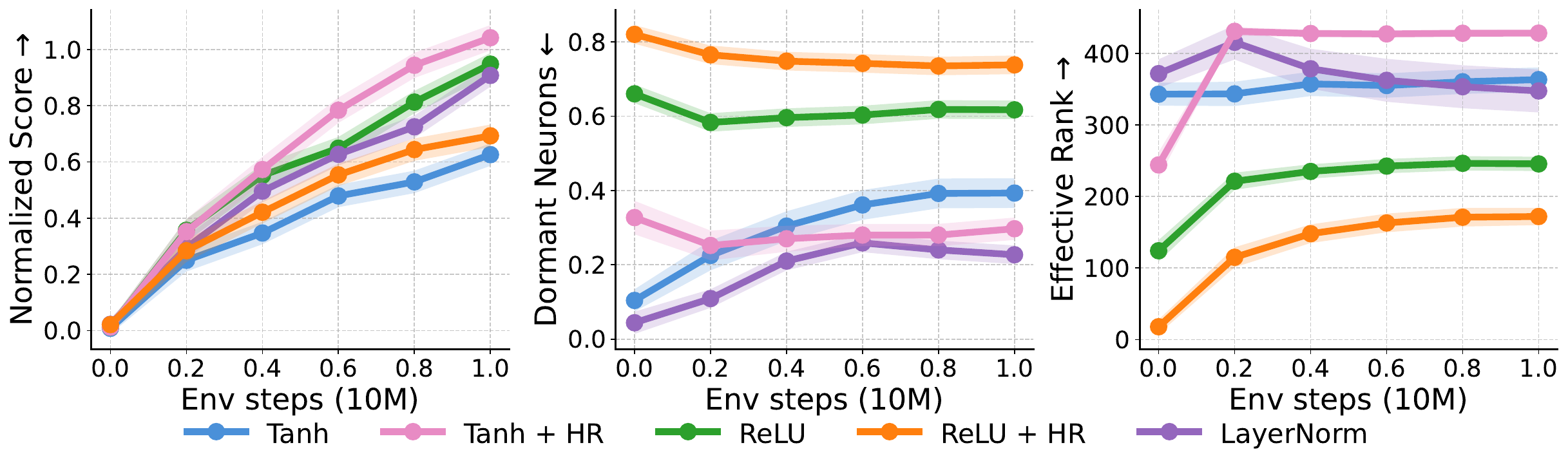}
    \caption{Training DQN on 8 Atari games for 10M iterations (40M frames) under five configurations. \textbf{(Left)} Aggregated normalized score. \textbf{(Middle)} Fraction of dormant neurons. \textbf{(Right)} Effective rank~\citep{Effective_rank_kumar_2019}. $\tanh$ activations exhibit strong dormancy comparable to $\relu$, while HR with $\tanh$ reduces dormancy and increases effective rank simultaneously.}
    \label{fig:dead_rank_dqn}
\end{figure}

We analyze DQN on 8 representative Atari games to understand what limits representation quality in deep RL, and use the resulting diagnosis to motivate our proposed architecture. \cref{fig:dead_rank_dqn} tracks performance, dormant neuron fraction, and effective rank of the representation $\mathbf{z}_t$ (here for DQN, we apply HR on the last hidden layer of the network) during training under five configurations: the original DQN architecture (\relu), DQN with $\tanh$, DQN with $\tanh$ + \texttt{LayerNorm}, and two variants with our proposed Hadamard Representation (HR), $\tanh$ + HR and \relu + HR.

\begin{figure*}[!htb]
    \centering
    \begin{subfigure}{0.5\textwidth}
        \centering
        \hspace{-4mm}
        \includegraphics[width=2.4in]{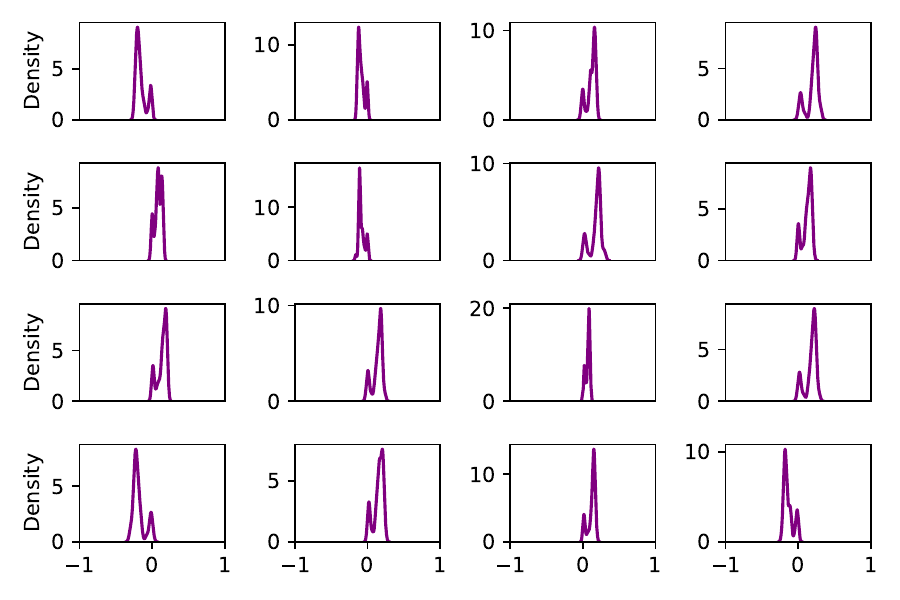}
        \caption{$\tanh$ at $10^{6}$ steps}
        \label{fig:tanh1m}
    \end{subfigure}%
    \begin{subfigure}{0.5\textwidth}
        \centering
        \includegraphics[width=2.4in]{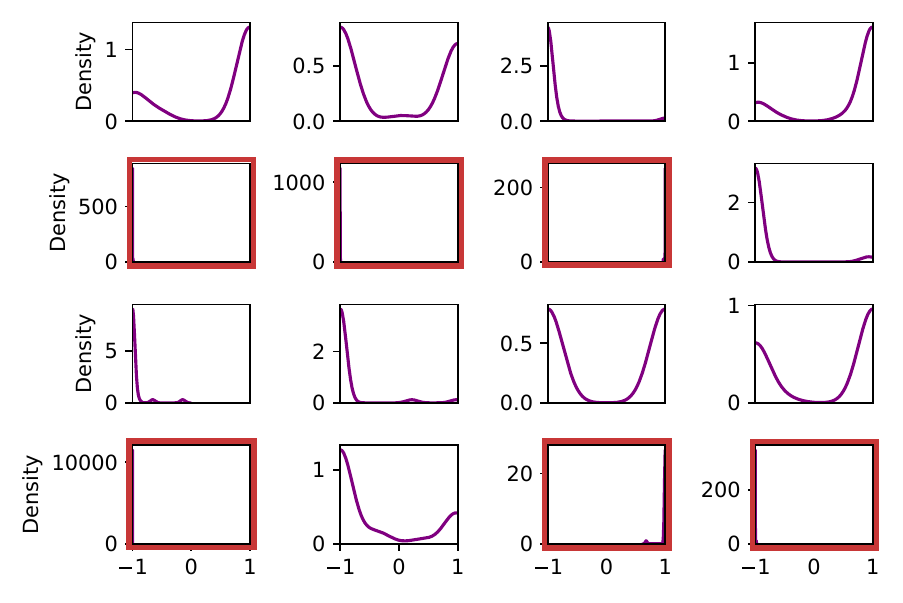}
        \caption{$\tanh$ at $10^{7}$ steps}
        \label{fig:tanh10m}
    \end{subfigure}
    \caption{Kernel density estimates (KDE) of 16 neurons from the compressed representation $\mathbf{z}_t \in \mathbb{R}^{512}$ during DQN training~\citep{Mnih2015Human-levelLearning} on Breakout with $\tanh$ activations. Red outlines mark dormant neurons whose density concentrates sharply at $\pm 1$, indicating a near-total loss of variance across observations.}
    \label{fig:kde_tanh_final}
\end{figure*}

\subsection{Revisiting Dormant Neurons in RL}

\paragraph{Observation 1: dormant $\relu$ costs capacity.}
The dormant $\relu$ phenomenon is well-studied in RL~\citep{PabloSamuel_Dying, dying_relu2, Tanh_Rank_Study_TMLR}. \cref{fig:dead_rank_dqn} confirms it: around 60\% of $\relu$ neurons become dormant during training, with correspondingly low effective rank. A natural goal follows: reduce dormant neurons to recover capacity.

\paragraph{Observation 2: $\tanh$ satisfies this goal and still does not win.}
$\tanh$ has fewer dormant neurons than $\relu$ ($\sim$40\% vs 60\%) and substantially higher effective rank, which is also observed by prior work~\citep{Tanh_Rank_Study_TMLR}. If the dormant count were the full story, $\tanh$ should outperform $\relu$. In our experiments it performs even worse. Something is missing.

\paragraph{Observation 3: dormant $\tanh$ neurons are not silent.}
The count understates the damage when the activation saturates to a nonzero value. For $\relu$ ($\Omega = 0$), a dormant neuron contributes zero and is effectively pruned. For $\tanh$ ($\Omega \in \{-1, +1\}$), it continues to contribute in a harmful way: turning weights into biases.

\begin{wrapfigure}[17]{r}{0.4\textwidth}
  \centering
  \includegraphics[width=0.38\textwidth]{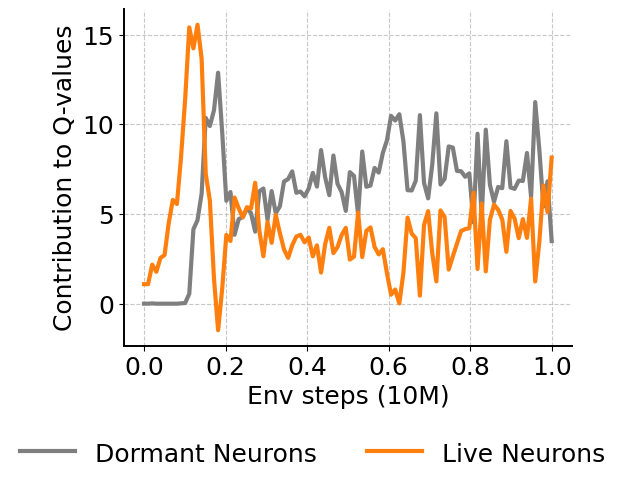}
  \caption{Average Q-value contribution of live and dormant neurons in the final hidden layer during 10M steps of DQN training with $\tanh$ activations on Seaquest. A dormant neuron retains a constant saturation value for any input.}
  \label{fig:hiddenbias}
\end{wrapfigure}

\begin{theorem}\label{theorem:bias}
When any subset of neurons in a hidden layer $\mathbf{z}^j$ collapses to a nonzero constant $\Omega$ across all observations, the pre-activation of the subsequent layer $\mathbf{h}^{j+1}$ effectively gains a fixed, input-independent bias $B^{j+1}_*$. The magnitude of this bias is determined by the outgoing weights $\mathbf{A}^{j+1}$ associated with the dormant neurons.
\end{theorem}

\begin{proof}
Let $\mathcal{D}$ be the subset of dormant neurons and $\mathcal{C}$ be the subset of active neurons in layer $j$. The pre-activation of the subsequent layer is defined as $\mathbf{h}^{j+1} = \mathbf{A}^{j+1} \mathbf{z}^j + \mathbf{b}^{j+1}$. Partitioning the input based on these sets yields:
\begin{equation}
    \mathbf{h}^{j+1} = \sum_{i \in \mathcal{C}} z^j_i \mathbf{A}^{j+1}_{:, i} + \underbrace{\sum_{i \in \mathcal{D}} \Omega \mathbf{A}^{j+1}_{:, i} + \mathbf{b}^{j+1}}_{\text{Effective Bias } B^{j+1}_{\text{eff}}},
\end{equation}
where $\mathbf{A}^{j+1}_{:, i}$ denotes the $i$-th column of the weight matrix. For $\relu$, $\Omega = 0$, meaning the second term vanishes and the layer is effectively pruned. For $\tanh$, $\Omega \in \{-1, +1\}$, so the dormant neurons contribute a constant vector $B^{j+1}_* = \sum_{i \in \mathcal{D}} \Omega \mathbf{A}^{j+1}_{:, i}$ that shifts the original bias $\mathbf{b}^{j+1}$. This shift persists as a fixed, input-independent distortion regardless of the input $\mathbf{x}$.
\end{proof}

\cref{fig:hiddenbias} illustrates this on Seaquest: If a neuron becomes dormant in the layer right before the output, it retains the same value for any input observation, but a multiplication of the nonzero saturation value with its outgoing weights implements a substantial ‘hidden’ bias on the Q-values. The Q-value contribution from dormant $\tanh$ neurons grows over training until it is comparable to that of live neurons, injecting a large state-independent offset into Q-value estimates. Figure~\ref{fig:kde_tanh_final} shows the corresponding activations progressively saturating toward $\pm 1$. For smooth activations, residual dormant neurons are actively harmful rather than merely inert, so an effective architectural intervention should prevent saturation in the first place.

\subsection{Hadamard Representation}
\label{sec:hr}

\begin{figure*}[!htb]
    \centering
    \includegraphics[width=0.83\textwidth]{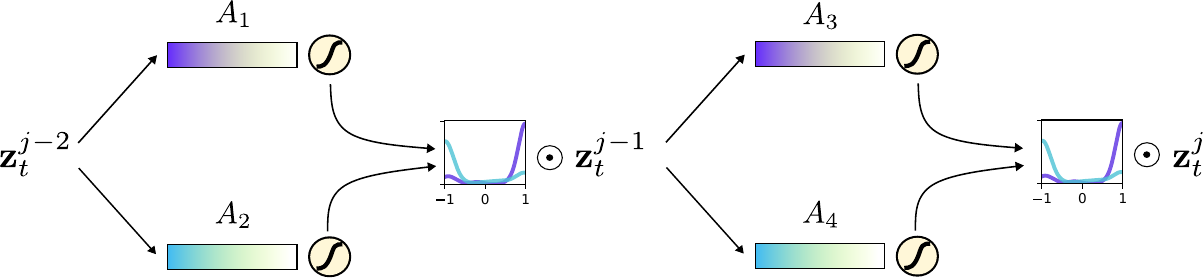}
    \vspace{4mm}
    \caption{The Hadamard representation. Horizontal bars represent weight vectors and $\mathbf{z}_t$ represents a hidden layer. Two parallel, independently parameterized activation layers are formed between consecutive hidden layers, and their element-wise product defines the propagated hidden layer.}
    \label{fig:Architecture}
\end{figure*}

A standard hidden layer is $\mathbf{z}(\mathbf{x}) = f(\mathbf{A}_1 \mathbf{x} + \mathbf{b}_1)$. HR augments this with a parallel, independently parameterized branch and defines the hidden layer as their element-wise product,
\begin{equation}
    \mathbf{z}(\mathbf{x}) = f(\mathbf{A}_1 \mathbf{x} + \mathbf{b}_1) \odot f(\mathbf{A}_2 \mathbf{x} + \mathbf{b}_2),
\end{equation}
which can be viewed as a highway layer with a closed carry gate~\citep{Highway} or as a symmetric variant of the Gated Linear Unit~\citep{Gated_Linear_Unit_Dauphin_Swiglu} in which both branches share the same activation. Figure~\ref{fig:Architecture} shows the architecture. HR improves representations through two complementary mechanisms.

\paragraph{Mechanism 1: saturation resistance.}
Our key hypothesis is that HR prevents saturation, thereby alleviating vanishing gradients and the emergence of dormant neurons. To support this, we examine the derivative of a product of two functions $g(x) \cdot h(x)$, which is defined as $g'(x)h(x) + g(x)h'(x)$. For a product of two $\tanh$ functions, $\mathbf{z}'(\mathbf{x})$ is:
\begin{equation*}
\mathbf{z}'(\mathbf{x}) = \mathbf{A}_1 \,\text{sech}^2(\mathbf{A}_1 \mathbf{x} + \mathbf{b}_1) \odot \tanh(\mathbf{A}_2 \mathbf{x} + \mathbf{b}_2) + \mathbf{A}_2 \,\text{sech}^2(\mathbf{A}_2 \mathbf{x} + \mathbf{b}_2) \odot \tanh(\mathbf{A}_1 \mathbf{x} + \mathbf{b}_1),
\end{equation*}
where $\text{sech}^2$ is the derivative of $\tanh$. Because $\tanh$ saturates to nonzero values ($\Omega \in \{-1, +1\}$), the product remains active unless both branches saturate simultaneously. Even if one branch saturates, the other preserves a non-trivial gradient path through its $\tanh$ factor, providing a mechanism to avoid vanishing gradients.

More formally, let $p$ denote the saturation probability of a single baseline neuron. Assuming branch independence and symmetric saturation probabilities, the probability of collapse for $\tanh$ reduces to $p^2$, since saturation of the element-wise product requires both independent branches to fail. Conversely, for $\relu$ where $\Omega = 0$, the product collapses if either branch saturates to zero; thus, the probability of collapse increases to $2p-p^2$ (the probability that one of the two neurons does not saturate is $1-p$, and the probability that both neurons do not saturate is $(1-p)^2$, thus probability that at least one neuron saturates is thus equal to $1-(1-p)^2=2p-p^2$). Table~\ref{tab:dying_neurons} shows that these theoretical shifts track empirical measurements across 8 Atari games ($-23\%$ for $\tanh$, $+18\%$ for $\relu$) closely. Furthermore, while \texttt{LayerNorm} reduces dormant $\tanh$ neurons more aggressively than HR, Figure~\ref{fig:dead_rank_dqn} demonstrates that HR produces a corresponding gain in effective rank, compared with the original $\tanh$.

\begin{table}[h]
\centering
\caption{Dormant neuron fractions under HR: theoretical predictions and empirical measurements averaged over 8 Atari games. HR reduces dormancy for $\tanh$ but amplifies it for ReLU, consistent with the predicted probabilities.}
\label{tab:dying_neurons}
\begin{tabular}{@{}lcccc@{}}
\toprule
\textbf{Activation} & \textbf{Without HR} & \textbf{With HR} & \textbf{$\Delta$ (predicted)} & \textbf{$\Delta$ (measured)} \\
\midrule
$\tanh$ & $p$ \hfill (0.39) & $p^2$ \hfill (0.30) & $-(p - p^2)$ & $-23\%$ \\
ReLU & $p$ \hfill (0.62) & $2p - p^2$ \hfill (0.73) & $+(p - p^2)$ & $+18\%$ \\
\bottomrule
\end{tabular}
\end{table}

\paragraph{Mechanism 2: multiplicative expressiveness.}
Independently of dormancy, each output neuron $z_i$ under HR computes a second-order interaction between two independently learned projections: $z_i(\mathbf{x}) = f(\mathbf{a}_{1, i}^\top \mathbf{x} + b_{1, i}) \cdot f(\mathbf{a}_{2, i}^\top \mathbf{x} + b_{2, i})$, where $\mathbf{a}_{k, i}^\top$ denotes the $i$-th row of the weight matrix $\mathbf{A}_k$. This architecture effectively expands the function class at a fixed layer width. Unlike the gating mechanisms in highway networks and GLUs, both branches in HR contribute activation mass equally rather than one branch acting as a scalar gate for the other. Section~\ref{sec:results_cc} demonstrates that this multiplicative structure produces effective rank gains even in regimes where dormant neurons are already rare, establishing it as an independent source of representational benefit.

\paragraph{When does each mechanism dominate?}
The two mechanisms are simultaneously active; their relative importance depends on the dormancy rate of the baseline. When dormant neuron rates are high (e.g., DQN and PPO on Atari), mechanism 1 is decisive and the activation must be $\tanh$: HR with $\relu$ amplifies dormancy and should be avoided. When dormant neuron rates are low (e.g., SimbaV2 and MR.Q, where LayerNorm is built into the architecture), mechanism 1 offers little to gain, mechanism 2 becomes the primary driver, and the original activation of the architecture can be preserved.

\section{Experiments}
\label{sec:exps}

\begin{figure}[!htb]
    \centering
    \includegraphics[width=0.9\linewidth]{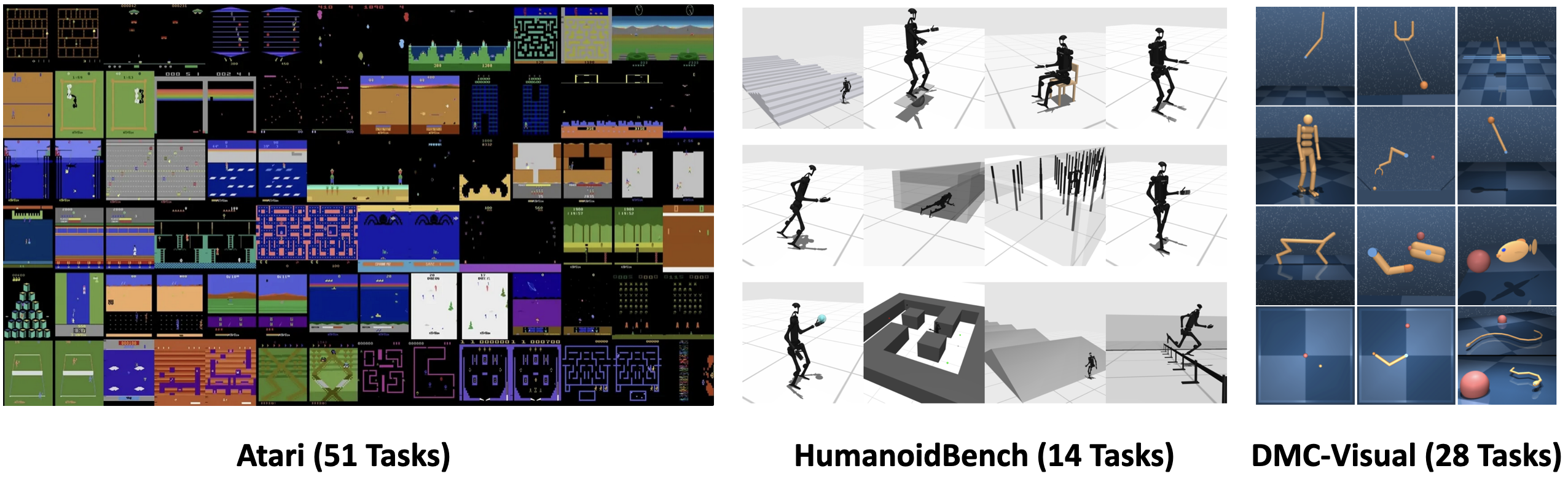}
    \caption{Three domains in our evaluation: pixel-based discrete control (Atari), state-based continuous control (HumanoidBench), and visual continuous control (DMC-Visual).}
    \label{fig:all_task}
\end{figure}

For all experiments, HR is applied \textbf{without} modifying or tuning any hyperparameters from the original baseline codebases. We evaluate HR across three domains selected to test the two mechanisms identified in Section~\ref{sec:method}: the high-dormancy regime (Atari), where saturation resistance should dominate, and the low-dormancy regime (state-based and visual continuous control), where multiplicative expressiveness serves as the primary driver. As detailed in the following sections, we confirm performance gains in the high-dormancy regime at scale (Section~\ref{sec:exp_atari}), isolate the expressiveness mechanism in the state-of-the-art continuous control methods (Section~\ref{sec:results_cc}), and rule out parameter count as a confounding factor via capacity-matched ablations (Section~\ref{sec:ablation}).
\paragraph{Algorithms and benchmarks.}
We evaluate HR across five algorithms spanning these regimes. For Atari, we utilize DQN~\citep{mnih2013playing} for qualitative representational analysis, PPO~\citep{schulman2017proximal} as a policy-gradient confirmation, and PQN~\citep{gallici2024simplifying} for a large-scale evaluation across 51 games. For continuous control, we pair HR with current state-of-the-art methods: SimbaV2~\citep{lee2025hyperspherical} on HumanoidBench and MR.Q~\citep{fujimoto2025towards} on DMC-Visual. All results are reported across 3--5 seeds per task. Aggregate scores are computed by first averaging over seeds per task and then calculating the Interquartile Mean (IQM) or median across the suite; normalization is applied per task prior to aggregation where noted. Detailed per-task comparisons are available in~\cref{app:eval_detail}.

\subsection{Atari: High-Dormancy Regime}
\label{sec:exp_atari}
\begin{wrapfigure}[16]{r}{0.7\textwidth}
  \centering
  \includegraphics[width=0.68\textwidth]{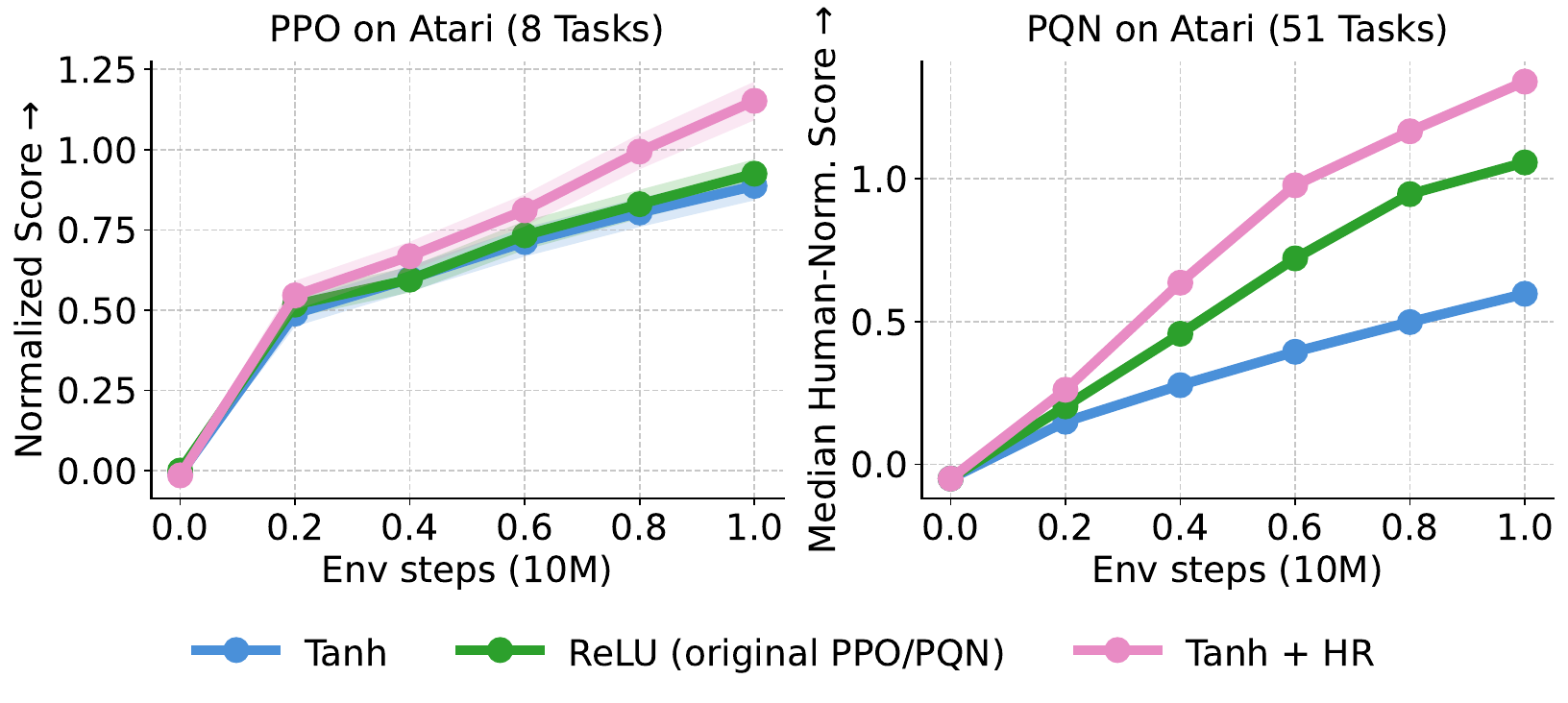}
  \caption{\textbf{Left:} Normalized performance on PPO across 8 Atari games. \textbf{Right:} Median human-normalized scores on PQN across 51 Atari games (5 seeds, 40M frames). HR applied to $\tanh$ surpass the standard $\tanh$ and $\relu$ baseline. No hyperparameters are changed for HR.}
  \label{fig:ppo_pqn}
\end{wrapfigure}
We confirm the DQN performance gain from Figure~\ref{fig:dead_rank_dqn} with two additional algorithms: PPO on 8 Atari tasks, and PQN on 51 Atari tasks at scale, each trained for 10M steps (40M frames).

\paragraph{PPO.}
In PPO, the final hidden layer $\mathbf{z}_t$ (where HR is applied on) feeds both the actor and the critic, and therefore receives gradients from both policy and value losses. After 40M frames, $\tanh$ + HR attains the best performance (\cref{fig:ppo_pqn}, left), while $\tanh$ and ReLU perform comparably. This mirrors the DQN results in~\cref{fig:dead_rank_dqn} and confirms that HR's gain generalizes across value-based and policy-gradient algorithms.

\paragraph{PQN at scale.}
PQN is a vectorized variant of DQN that delivers both faster training and improved convergence~\citep{pqn}. We apply HR in all hidden layers and evaluate across 51 non-exploration-driven Atari games with 5 seeds per game. \cref{fig:ppo_pqn} (right) shows that HR increases the median human-normalized score substantially, with $\tanh$ + HR improving over standard $\tanh$ by more than 100\%. $\relu$ outperforms $\tanh$ here, as it does in DQN, which is consistent with the broader observation that continuously differentiable activations are generally disfavored relative to $\relu$ in RL~\citep{Teney_2024_CVPR_redshift}; HR closes this gap while preserving the high effective rank that $\tanh$ naturally supports. We also show that HR is more effective than other methods that are proposed to prevent the loss plasticity in RL (i.e. CReLU~\citep{CReLU_continual_zaheerabbas} and ReDo~\citep{PabloSamuel_Dying}, see~\cref{app:redo}).

\subsection{Continuous Control: Low-Dormancy Regime}
\label{sec:results_cc}

Beyond Atari, we evaluate HR on two continuous control settings: 14 tasks from HumanoidBench (low-dimensional state-based) and 7 tasks from DMC-Visual-Hard (pixel-based); results on the full 28-task DMC-Visual suite are reported in Appendix~\ref{app:dmc_visual}. In each setting we augment the current state-of-the-art method with HR: SimbaV2 (HR is applied on on the first hidden layer of the SimbaV2 block) on HumanoidBench and MR.Q (HR is applied on the encoder) on DMC-Visual. Both baselines already exhibit negligible dormant neuron rates, which isolates mechanism 2 (multiplicative expressiveness) as the only one available for HR to leverage.

\begin{figure}[!htb]
    \centering
    \includegraphics[width=0.9\linewidth]{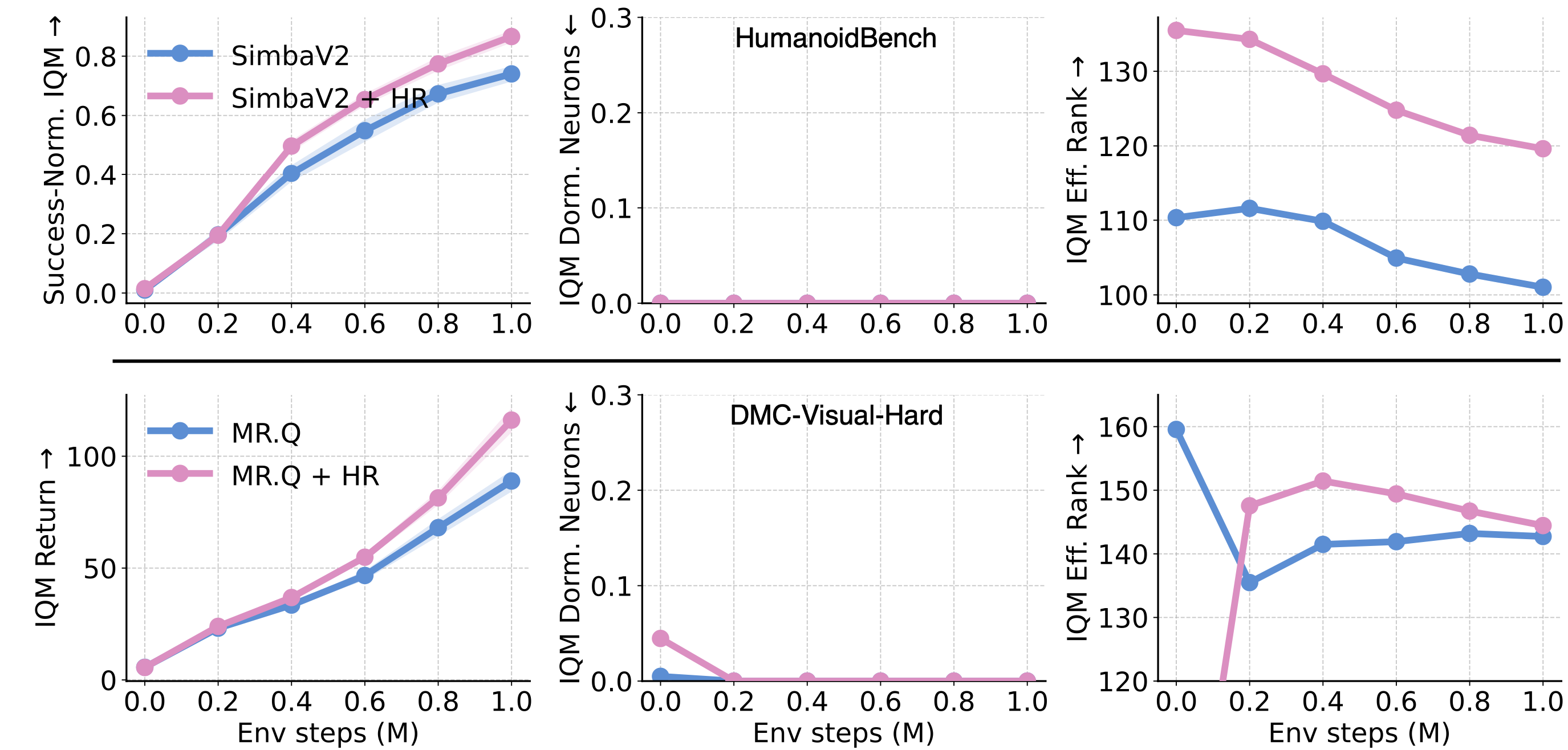}
    \caption{IQM Aggregated performance, dormant neuron fraction, and effective rank on 14 HumanoidBench tasks (top) and 7 DMC-Visual-Hard tasks (bottom). Dormant neurons are already negligible without HR in both settings, yet HR consistently improves effective rank and downstream performance, isolating multiplicative expressiveness as an independent source of gain.}
    \label{fig:cc}
\end{figure}

Figure~\ref{fig:cc} shows that despite near-zero dormant neuron rates in both baselines, HR increases the effective rank and yields corresponding performance improvements. This confirms that HR's gains do not depend solely on the presence of dormant neurons, and that its multiplicative expressiveness is an independent source of benefit. Combined with the Atari results, these findings support the two-mechanism account: HR's primary contribution shifts from saturation resistance (Atari, $\tanh$-specific) to multiplicative expressiveness (continuous control, activation-agnostic) as the dormancy rate of the baseline decreases.

\begin{figure}[!htb]
    \centering
    \includegraphics[width=1\linewidth]{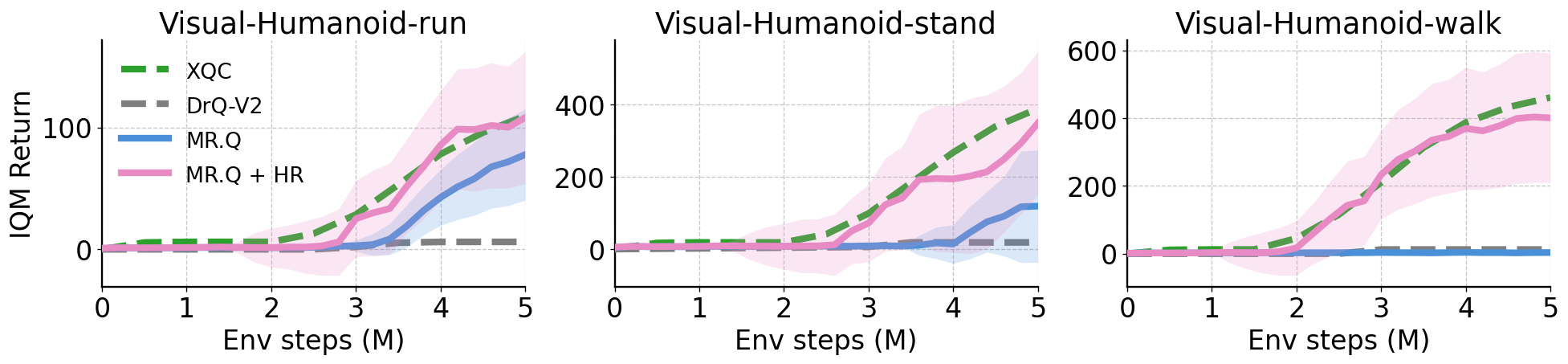}
    \caption{Evaluation on DMC-Visual-Humanoid tasks. MR.Q + HR matches XQC and substantially outperforms MR.Q and DrQ-v2. Results for DrQ-v2 and XQC are taken from~\citep{palenicek2025xqc}.}
    \label{fig:humanoid}
\end{figure}

\paragraph{Extended training on Visual Humanoid tasks.}
The three DMC-Visual-Humanoid tasks are particularly slow to converge, which makes the standard training budget insufficient to clearly differentiate methods. To obtain a stronger learning signal on these tasks, we extend training to $5\times$ and compare against two strong baselines: DrQ-v2~\citep{yarats2021mastering} and the recent state-of-the-art XQC~\citep{palenicek2025xqc} for DMC-Visual-Humanoid tasks. As shown in Figure~\ref{fig:humanoid}, MR.Q + HR outperforms the MR.Q baseline and performs on par with XQC, while DrQ-v2 fails to learn within the 5M-step budget. DrQ-v2 and XQC results are taken from the XQC paper.

\subsection{Ablations}
\label{sec:ablation}

We conduct two ablation experiments for HR on 8 Atari tasks (DQN), 14 HumanoidBench tasks (SimbaV2), and 7 DMC-Visual-Hard tasks (MR.Q).

\paragraph{Does HR just add parameters?}
HR maintains two parallel branches and therefore increases the parameter count of the hidden layer. To check whether the gains come from this added capacity rather than from the multiplicative structure, we compare HR to a parameter-matched wider baseline for each algorithm: we double the DQN hidden layer to match the $2\times$ weight count of $\tanh$ + HR, and analogously widen SimbaV2 and MR.Q to match their respective HR variants. Table~\ref{tab:hr_vs_widen} shows that naive widening does not consistently improve performance and, in the DQN case, substantially degrades it, which is also shown by recent work~\citep{Pruned_network_rl, mixture_of_experts}. HR, in contrast, yields gains across all three algorithms. Parameter count alone does not explain HR's benefits.

\begin{table}[!htb]
\centering
\caption{HR versus parameter-matched widening across DQN, SimbaV2, and MR.Q. HR outperforms both the baseline and the widened variant in all three settings.}
\label{tab:hr_vs_widen}
\begin{tabular}{@{}lccc@{}}
\toprule
\textbf{Variant} & \textbf{DQN} & \textbf{SimbaV2} & \textbf{MR.Q} \\
\midrule
Baseline & 0.91 & 0.71 & 1.00 \\
+ HR     & \cellcolor{green!25}1.03 {\scriptsize ($+13\%$)} & \cellcolor{green!25}0.84 {\scriptsize ($+18\%$)} & \cellcolor{green!25}1.17 {\scriptsize ($+17\%$)} \\
+ Widen  & \cellcolor{red!30}0.49 {\scriptsize ($-46\%$)}    & \cellcolor{green!12}0.76 {\scriptsize ($+7\%$)}  & \cellcolor{red!10}0.97 {\scriptsize ($-3\%$)} \\
+ 2HR    & \cellcolor{green!22}1.01 {\scriptsize ($+12\%$)} & --- & --- \\
\bottomrule
\end{tabular}
\end{table}

\paragraph{Does stacking HR help?}
We investigate whether stacking HR modules yields further gains by applying two HR blocks in DQN (2HR). Table~\ref{tab:hr_vs_widen} shows that 2HR performs comparably to a single HR: both improve over the baseline by a similar margin, with no additional benefit from a second HR block.

\section{Conclusion and Discussion}
\label{sec:conclusion}
We analyzed representational failure modes in deep RL, demonstrating that saturated $\tanh$ neurons do not merely stop learning but actively corrupt downstream layers by converting outgoing weights into fixed, silent biases. To address this, we proposed the Hadamard Representation (HR), a lightweight architectural modification that replaces standard hidden layers with the element-wise product of two independent activation branches. HR improves representational quality through two distinct mechanisms: saturation resistance, which quadratically reduces the probability of $\tanh$ collapse, and multiplicative expressiveness, which increases effective rank independently of dormancy. Evaluated across five algorithms and three diverse domains, HR consistently enhances performance, rank, and dormant rates without hyperparameter tuning, outperforming capacity-matched wider baselines.

\paragraph{Limitations and future work.}
As we maintained baseline hyperparameters throughout this work, dedicated tuning for HR remains a clear path for further gains. While we evaluated HR on a representative set of algorithms, future research could explore its integration into model-based frameworks~\citep{Hafner2020MasteringModels,hansen2024tdmpc2} or its interaction with specialized auxiliary losses~\citep{Schwarzer2020Data-EfficientRepresentations}. More broadly, our findings indicate that the potential of non-piecewise-linear activations in RL is significantly underestimated. Extending this study to supervised and continual learning settings~\citep{CReLU_continual_zaheerabbas, delfosse2024adaptive} offers a promising direction for future architectural innovation. We hope HR inspires future research on network architectural design in RL.


\bibliography{main}
\bibliographystyle{abbrv}

\newpage

\appendix
\addcontentsline{toc}{section}{Appendix}
\part{Appendix}
\parttoc
\clearpage

\section{Impact Statement}
\label{app:is}
This research demonstrates that architectural advancements, such as Hadamard Representation, yield substantial performance gains in RL. By enhancing the efficiency and accessibility of AI, these innovations catalyze broader adoption and facilitate the exploration of learning systems across a wide range of real-world applications.

\section{Implementation Details}\label{app:Implementation}

\subsection{Hyperparameters}\label{app:Hyperparameters}

For DQN and PPO, we evaluate on 8 Atari environments using 5 random seeds. For the mean scores, we take the mean over the eight environments. Our baseline-normalized score is calculated with respect to the original implementation using a ReLU activation. All hyperparameters used in our experiments for DQN, PPO, and PQN follow \verb|cleanrl| \citep{huang2022cleanrl} and the original PQN release \citep{pqn}; they are reproduced in Tables~\ref{tab:hyperparametersdqn}, \ref{tab:hyperparametersppo}, and~\ref{tab:hyperparameterspqn}. 
For continuous control baselines, we use the official codebase for Simbav2~\footnote{SimbaV2: \url{https://github.com/DAVIAN-Robotics/SimbaV2}} and MR.Q~\footnote{MR.Q: \url{https://github.com/facebookresearch/MRQ}} and their default configurations without touching anything. For XQC and DrQ-v2 results in~\cref{fig:humanoid}, we obtain them from the published XQC~\citep{palenicek2025xqc} paper.

\begin{table}[htbp]
\centering
\caption{DQN Hyperparameters}
\label{tab:hyperparametersdqn}
\begin{tabular}{@{}lll@{}}
\toprule
\textbf{Hyperparameter} & \textbf{Value} & \textbf{Description} \\ \midrule
Learning Rate & $1 \times 10^{-4}$ & Learning rate for the optimizer \\
Discount Factor $(\gamma)$ & 0.99 & Discount for future rewards \\
Replay Memory Size & 1,000,000 & Size of the experience replay buffer \\
Mini-batch Size & 32 & Number of samples per batch update \\
Target Network Update Frequency & 1000 & Update frequency for the target network \\
Initial Exploration & 1.0 & Initial exploration rate in $\epsilon$-greedy \\
Final Exploration & 0.1 & Final exploration rate in $\epsilon$-greedy \\
Final Exploration Frame & 1,000,000 & Frame number to reach final exploration \\
Exploration Decay Frame & 1,000,000 & Frames over which exploration rate decays \\
Action Repeat (Frame Skip) & 4 & Number of frames skipped per action \\
Reward Clipping & [-1, 1] & Range to which rewards are clipped \\
Input Dimension & 84 x 84 & Dimensions of the input image \\
Latent Dimension & 512 & Dimension of the latent representation \\
Input Frames & 4 & Number of frames used as input \\
Training Start Frame & 80,000 & Frame number to start training \\
Loss Function & Mean Squared Error & Loss function used for updates \\
Optimizer & Adam & Optimization algorithm used \\
Optimizer $\epsilon$ & $10^{-5}$ & Adam Epsilon \\
\bottomrule
\end{tabular}
\end{table}

\begin{table}[htbp]
\centering
\caption{PPO Hyperparameters}
\label{tab:hyperparametersppo}
\begin{tabular}{@{}lll@{}}
\toprule
\textbf{Hyperparameter} & \textbf{Value} & \textbf{Description} \\ \midrule
Learning Rate & $2.5 \times 10^{-4}$ & Learning rate for the optimizer \\
Discount Factor $(\gamma)$ & 0.99 & Discount factor for future rewards \\
Number of Steps & 128 & Number of steps per environment before update \\
Anneal LR & True & Whether to anneal the learning rate \\
GAE Lambda & 0.95 & Lambda parameter for GAE \\
Number of Minibatches & 4 & Number of minibatches to split the data \\
Update Epochs & 4 & Number of epochs per update \\
Normalize Advantage & True & Whether to normalize advantage estimates \\
Clipping Coefficient & 0.1 & Clipping parameter for PPO \\
Clip Value Loss & True & Whether to clip value loss \\
Entropy Coefficient & 0.01 & Coefficient for entropy bonus \\
Value Function Coefficient & 0.5 & Coefficient for value function loss \\
Maximum Gradient Norm & 0.5 & Maximum norm for gradient clipping \\
Target KL & None & Target KL divergence between updates \\
Latent Dimension & 512 & Dimension of the latent representation \\
Optimizer & Adam & Optimization algorithm used \\
Optimizer $\epsilon$ & $10^{-5}$ & Adam Epsilon \\
\bottomrule
\end{tabular}
\end{table}

\begin{table}[!t]
\centering
\caption{PQN Hyperparameters \citep{pqn}}
\label{tab:hyperparameterspqn}
\begin{tabular}{@{}lll@{}}
\toprule
\textbf{Hyperparameter} & \textbf{Value} & \textbf{Description} \\ \midrule
Total Timesteps & 10,000,000 & Total timesteps for training \\
Timesteps for Decay & 10,000,000 & Timesteps for decay functions (epsilon and lr) \\
Number of Environments & 128 & Number of parallel environments \\
Steps per Environment & 32 & Steps per environment in each update \\
Number of Epochs & 2 & Number of epochs per update \\
Number of Minibatches & 32 & Number of minibatches per epoch \\
Starting Epsilon & 1.0 & Starting epsilon for exploration \\
Final Epsilon & 0.001 & Final epsilon for exploration \\
Epsilon Decay Ratio & 0.1 & Decay ratio for epsilon \\
Epsilon for Test Policy & 0.0 & Epsilon for greedy test policy \\
Learning Rate & 0.00025 & Learning rate \\
Learning Rate Linear Decay & True & Use linear decay for learning rate \\
Max Gradient Norm & 10.0 & Max gradient norm for clipping \\
Discount Factor $(\gamma)$ & 0.99 & Discount factor for reward \\
Lambda $(\lambda)$ & 0.65 & Lambda for generalized advantage estimation \\
Episodic Life & True & Terminate episode when life is lost \\
Reward Clipping & True & Clip rewards to range [-1, 1] \\
Frame Skip & 4 & Number of frames to skip \\
Max No-Ops on Reset & 30 & Max number of no-ops on reset \\
Test During Training & True & Run evaluation during training \\
Number of Test Envs & 8 & Number of environments used for testing \\
\bottomrule
\end{tabular}
\end{table}

\subsection{Hadamard Implementation}\label{app:hadamard}

\begin{figure}[!htb]
    \centering
    \includegraphics[width=0.8\linewidth]{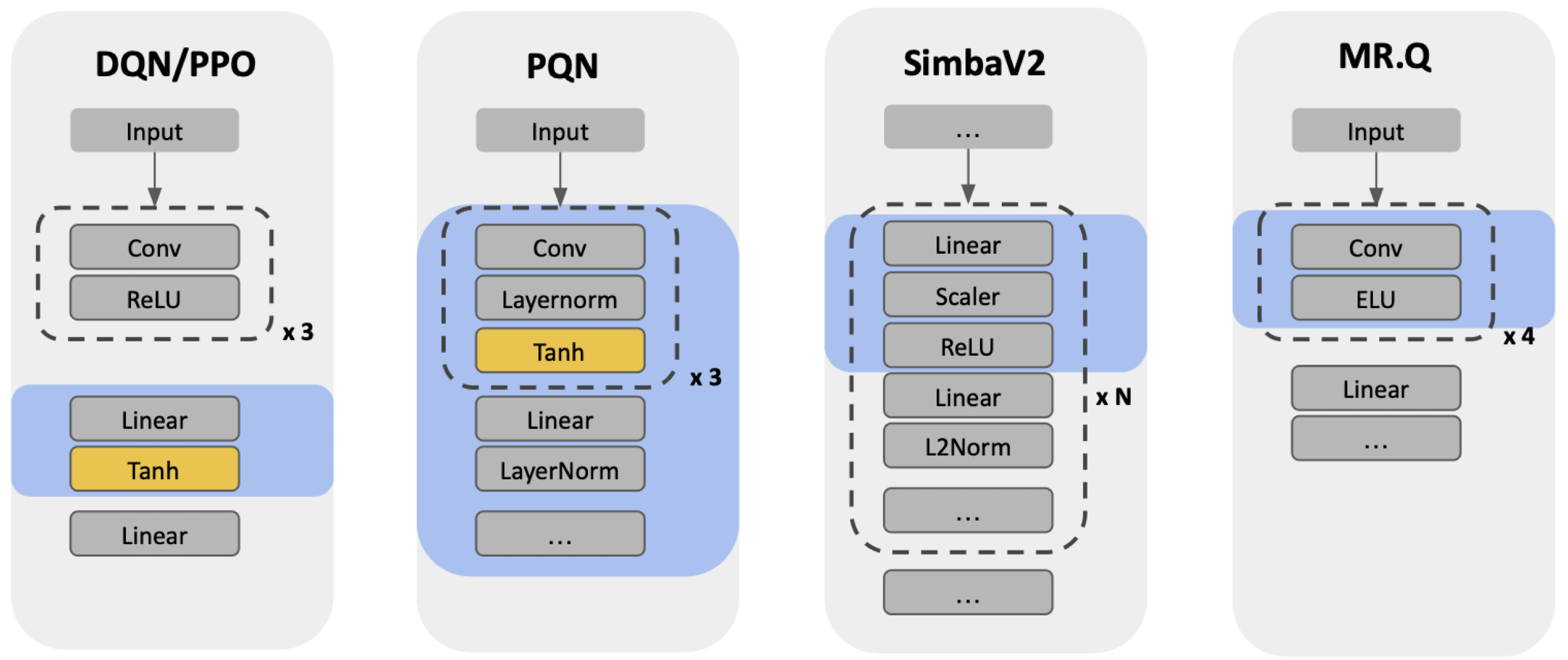}
    \caption{Architectural diagrams highlighting (in blue) the locations where the HR structure is integrated within the different baseline methods.}
    \label{fig:hr_where}
\end{figure}

The Hadamard Representation (HR) is a general architectural modification that is agnostic to the specific type of layer used. It is constructed by creating two parallel branches—each processing the same input—and computing the element-wise product of their respective non-linear activations. While the integration points vary by baseline (visualized in \cref{fig:hr_where}), the fundamental requirement is simply an additional set of weights to form the second branch.

While any layer type (e.g., convolutional or linear) can be used, we illustrate the implementation using a standard linear layer as a representative example:

\begin{verbatim}
# hidden: input from the preceding layer (e.g., encoder output)
# Initialize two parallel layers (example using Linear)
self.branch1 = nn.Linear(input_dim, output_dim)
self.branch2 = nn.Linear(input_dim, output_dim)

# Compute parallel activations using a non-linearity (e.g., Tanh)
out1 = torch.tanh(self.branch1(hidden))
out2 = torch.tanh(self.branch2(hidden))

# Hadamard product (element-wise multiplication)
hr_output = out1 * out2
\end{verbatim}

\clearpage
\section{Experimental Details}

\subsection{KDE Implementation Details}
\label{app:kde}
To stabilize KDE computation and avoid singularity issues, a small noise $\epsilon$ following a normal distribution is added to each neuron's activations: $\alpha'_i = \alpha_i + \epsilon$, with $\epsilon \sim \mathcal{N}(0, \sigma^2)$ and $\sigma^2 = 1 \times 10^{-5}$. The bandwidth, crucial for accuracy, is computed using Scott's rule, adjusted by the standard deviation of the jittered activations: $bw = n^{-1/5} \cdot \text{std}(\alpha'_i)$, where $n$ is the number of samples in $\alpha_i$. The density is estimated using a Gaussian kernel, $f(x) = \frac{1}{n \cdot bw} \sum_{j=1}^n K\!\left(\frac{x - \alpha'_{ij}}{bw}\right)$. A neuron is classified as dormant when $\max(f(x)) \geq \omega$. After analyzing individual KDEs, $\omega = 20$ provides a strong approximation of actual dormant neurons.

\subsection{Effective Rank Implementation}\label{app:rank_calc}
\begin{verbatim}
def compute_rank_from_features(feature_matrix, rank_delta=0.01):
    sing_values = np.linalg.svd(feature_matrix, compute_uv=False)
    cumsum = np.cumsum(sing_values)
    nuclear_norm = np.sum(sing_values)
    approximate_rank_threshold = 1.0 - rank_delta
    threshold_crossed = (cumsum >= approximate_rank_threshold * nuclear_norm)
    effective_rank = sing_values.shape[0] - np.sum(threshold_crossed) + 1
    return effective_rank
\end{verbatim}

\newpage
\subsection{Increasing Representation Parameters}\label{app:parameters}
\begin{figure}[!htb]
    \begin{subfigure}{0.49\textwidth}
        \centering
        \includegraphics[width=2in]{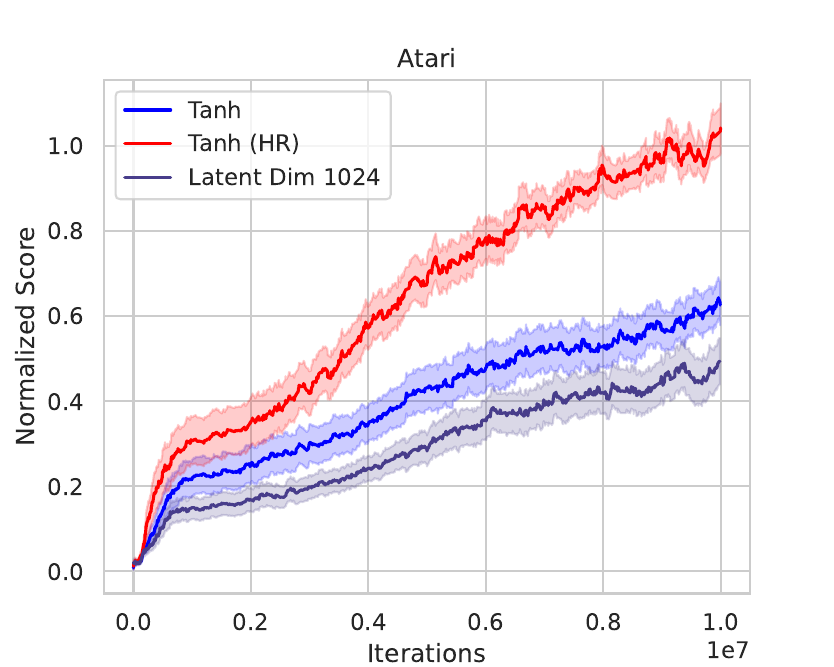}
        \caption{Performance}
    \end{subfigure}
    \begin{subfigure}{0.49\textwidth}
        \centering
        \includegraphics[width=2in]{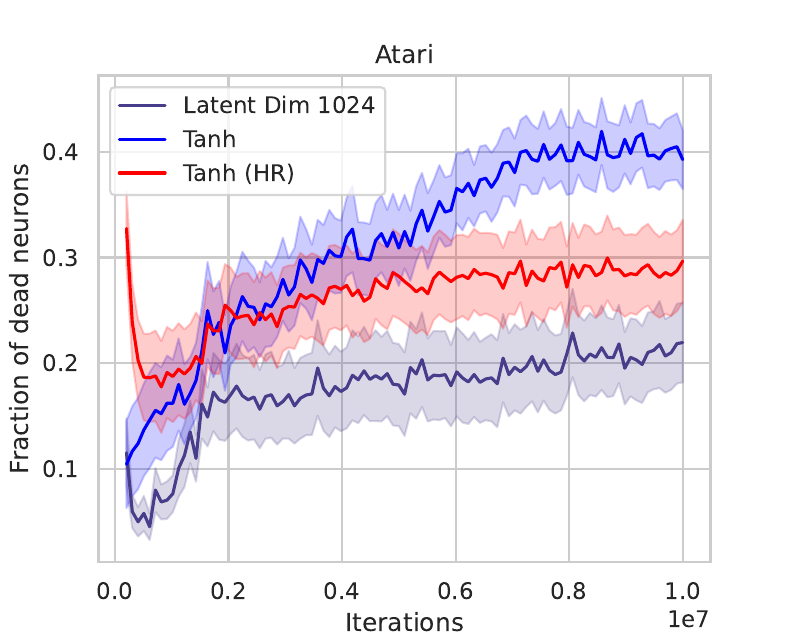}
        \caption{Dormant neurons}
    \end{subfigure}
    \begin{subfigure}{0.49\textwidth}
        \centering
        \includegraphics[width=2in]{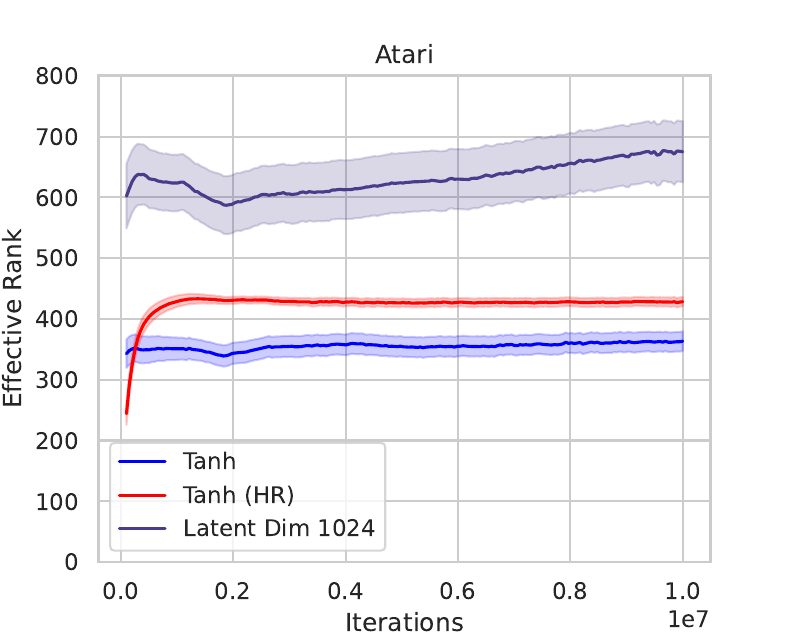}
        \caption{Effective rank}
    \end{subfigure}
    \begin{subfigure}{0.49\textwidth}
        \centering
        \includegraphics[width=2in]{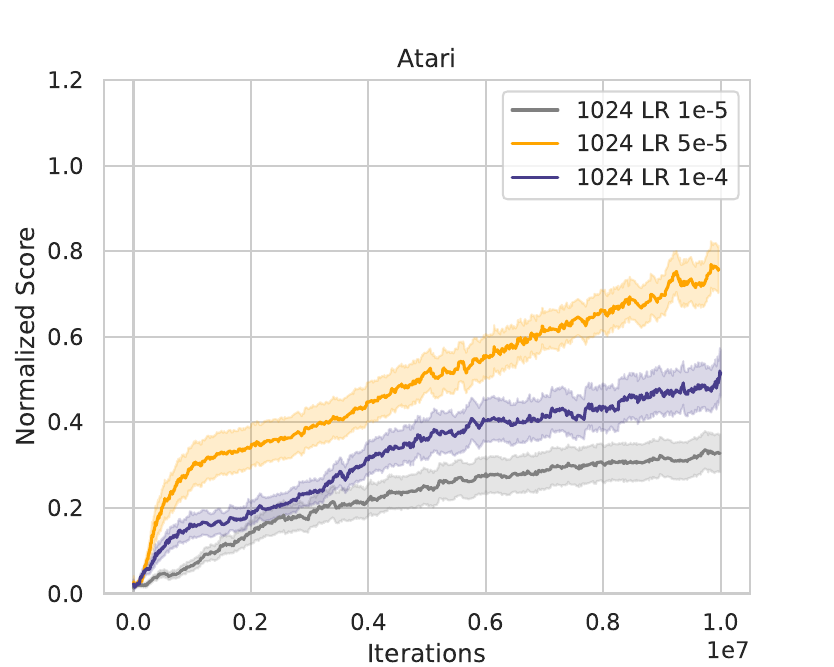}
        \caption{Learning rates, $z_{t} \in \mathbb{R}^{1024}$}
    \end{subfigure}
    \caption{Comparison of standard tanh, tanh with a higher representation dimension $\z_{t} \in \mathbb{R}^{512 \rightarrow 1024}$, and a tanh Hadamard representation, on performance (a), fraction of dormant neurons (b), effective rank of $\z_{t}$ (c), and learning rates of the higher-dimensional latent (d). Increasing the representation dimension naturally raises the effective rank, but a larger representation dimension is not always preferable: it often requires different hyperparameters and can reduce performance \citep{Pruned_network_rl, mixture_of_experts}. The learning-rate ablation in (d) shows that wider layers can prefer lower learning rates, but HR still significantly outperforms any of the 1024-dimensional learning-rate ablations, indicating that HR's improvement is not explained by parameter count.}
    \label{fig:1024}
\end{figure}

\newpage
\subsection{Compare HR with CReLU and ReDo}
\label{app:redo}

We compare HR against two baselines that are proposed to prevent the loss of plasticity in RL. CReLU~\citep{CReLU_continual_zaheerabbas} focuses on architectural robustness by preserving both positive and negative signal components to ensure continuous gradient flow. ReDo~\citep{PabloSamuel_Dying}, on the other hand, employs an active recycling strategy that identifies and resets dormant neurons based on an activation threshold $\tau$. 

\begin{figure}[!htb]
    \centering
    \includegraphics[width=0.5\linewidth]{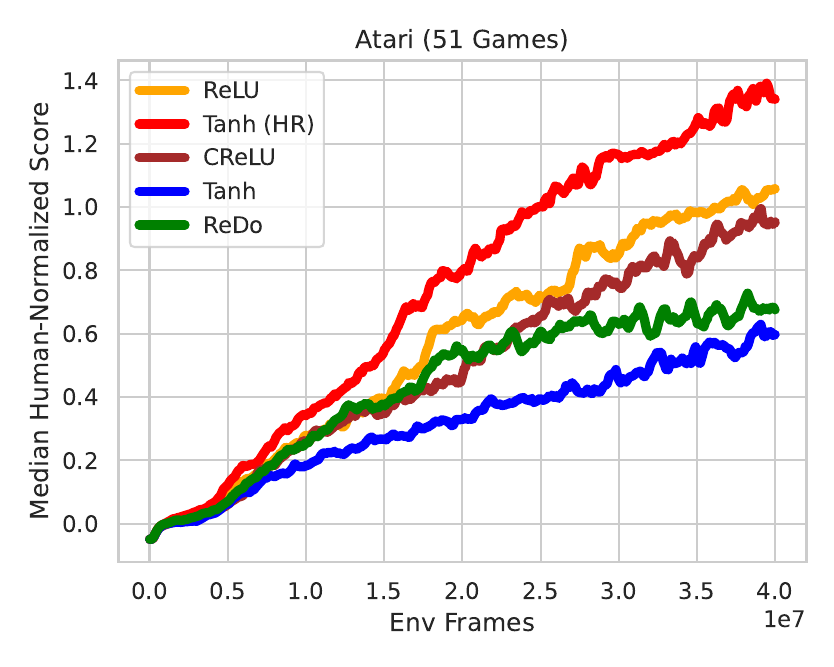}
    \caption{Median Human-Normalized scores on PQN in 51 Atari games for 5 seeds over 40M frames. Labels represent encoder activations. Employing a HR provides a significant performance improvement over the recent PQN baseline (displayed as ReLU), the CReLU~\citep{CReLU_continual_zaheerabbas} and ReDo ($\tau$=0.025)~\citep{PabloSamuel_Dying}.}
    \label{fig:51_all}
\end{figure}

As illustrated in Figure~\ref{fig:51_all}, HR provides a substantial performance improvement over both the recent PQN baseline (labeled as ReLU) and these specialized plasticity-preservation methods.

\subsection{Training Details}
\label{app:gpu}
We train all experiments on a cluster of A100 GPUs. Adding HR slows down the training, roughly 1.2x.

\section{Evaluation Details and Individual Task Performance}
\label{app:eval_detail}
\subsection{DQN \& PPO on 8 Atari}\label{app:atari_eval_details}
\begin{figure}[!htb]
    \centering
    \includegraphics[width=0.5 \textwidth]{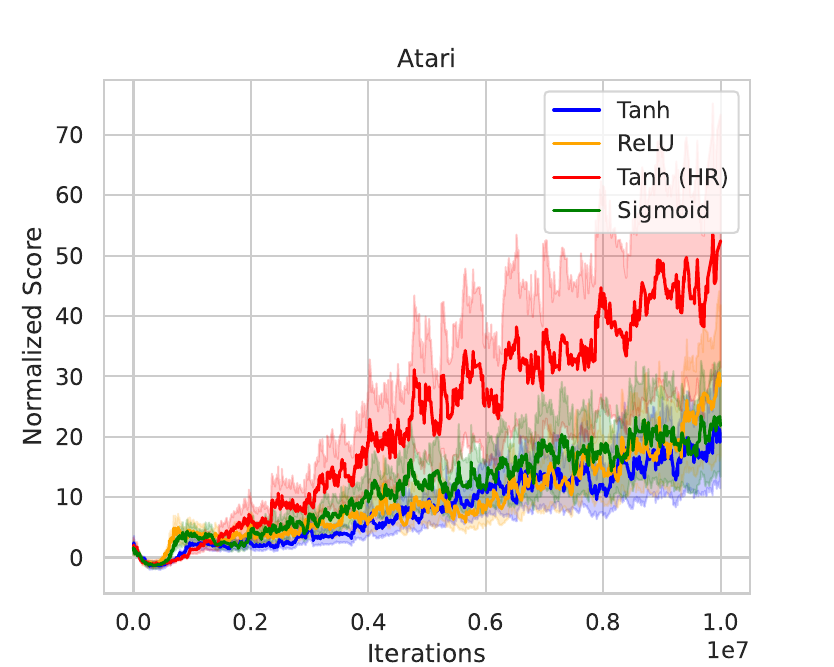}
    \caption{Human-Normalized performance (in multiples) with the standard deviation over the means in the Atari domain for 10M iterations (40M frames).}
    \label{fig:human_normalized}
\end{figure}

For DQN and PPO the Hadamard representation was applied to the final hidden layer of the Nature CNN. Performance is normalized with respect to the ReLU baseline on which the experiments were built \citep{huang2022cleanrl}. The minimum and maximum scores of the ReLU baseline are taken for each environment, and the normalized score is
\begin{equation}
    \text{Normalized Score} = \frac{\text{Score} - \text{Min Score}}{\text{Max Score} - \text{Min Score}},
\end{equation}
where \textit{Min Score} and \textit{Max Score} are the lowest and highest scores recorded by the ReLU baseline (the lowest is typically the random-policy score). To average, we sum the normalized scores for every run and take the mean. The Human-Normalized Score \citep{Mnih2015Human-levelLearning} is computed analogously using human and random performance:
\begin{equation}
    \text{Human-Normalized Score} = \frac{\text{Score} - \text{Random Score}}{\text{Human Score} - \text{Random Score}}.
\end{equation}
Computing performance under this normalization gives Fig.~\ref{fig:human_normalized}. Because we evaluate on a subset of Atari for DQN and PPO, the VideoPinball environment dominates the Human-Normalized Score; for fairer comparison we therefore use baseline-normalized scores in the main paper.

\begin{figure}[!htb]
    \centering
    \includegraphics[width=0.6\textwidth]{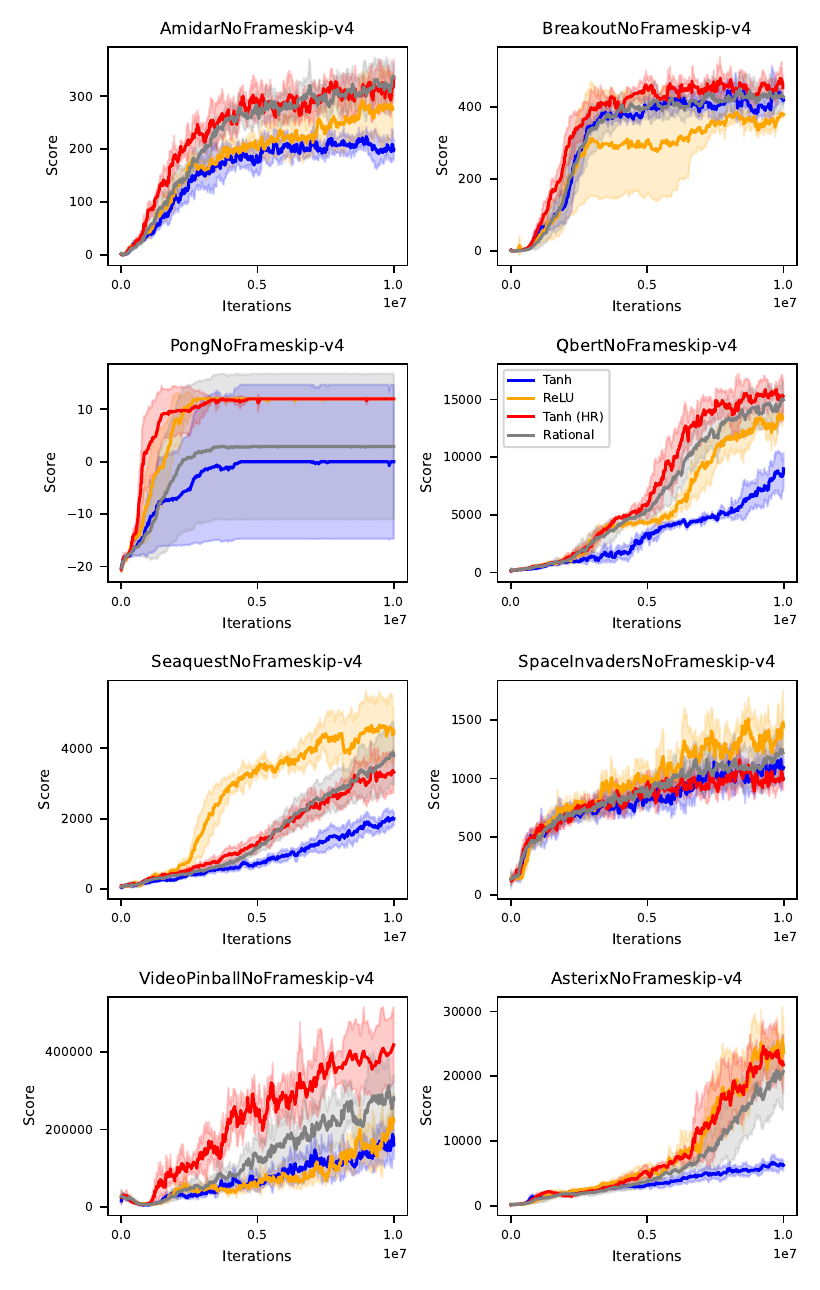}
    \caption{DQN per-environment performance. Lines are means over 5 seeds, shaded regions show standard deviation.}
\end{figure}

\begin{figure}[!htb]
    \centering
    \includegraphics[width=0.64\textwidth]{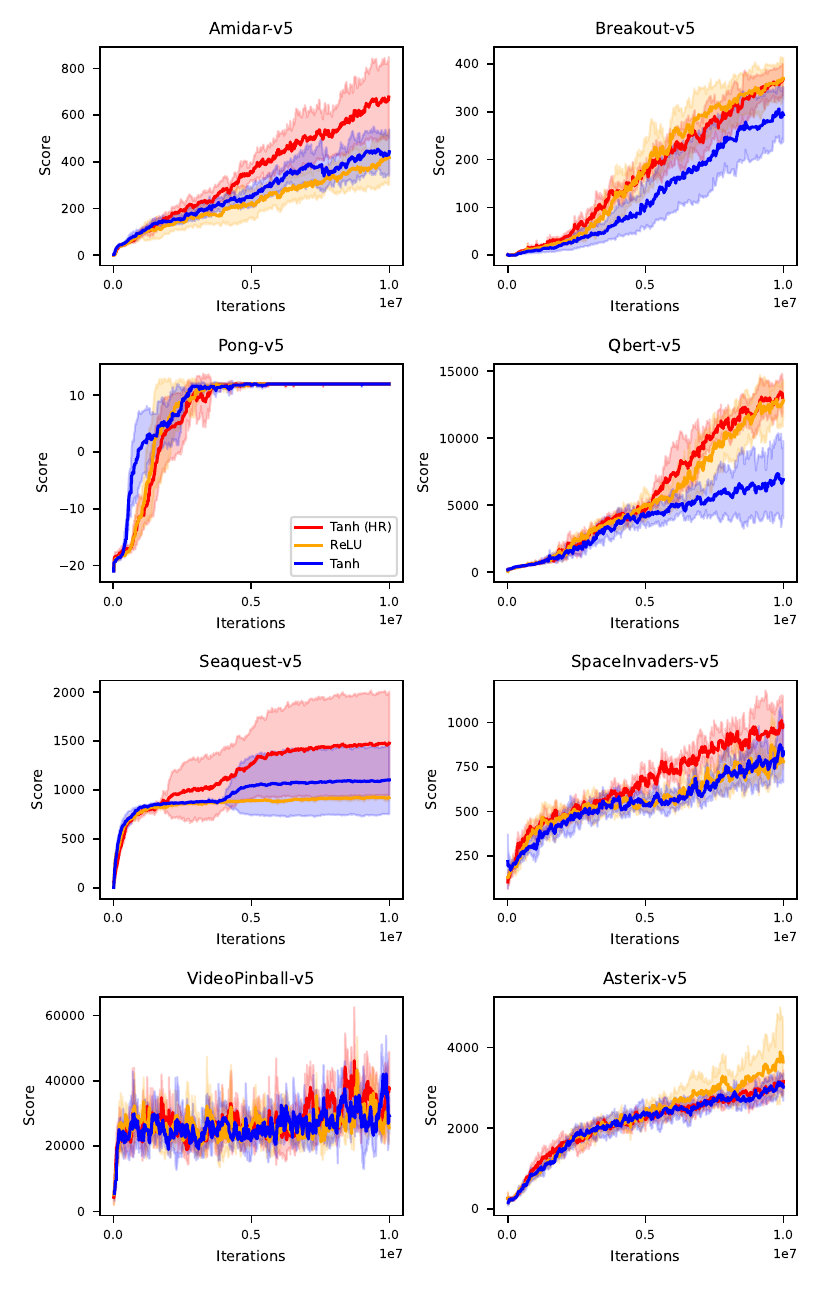}
    \caption{PPO per-environment performance. Lines are means over 5 seeds, shaded regions show standard deviation.}
    \label{fig:ppo_indiviudal1}
\end{figure}

\subsection{PQN on 51 Atari}
We use Median Human-Normalized Score for PQN. For each game, compute the average score $x_i$ across multiple independent seeds. Then compute the normalized score $Z_i$ as:
\begin{equation}
    Z_i = \frac{x_i - r_i}{h_i - r_i}
\end{equation}
where $x_i$ is the raw score, and $r_i$ and $h_i$ are the random and human scores for game $i$ (we take these values from~\citep{kooi2025hadamax}), respectively. After computing the normalized scores for all 51 games * seeds, they are sorted and the median value is computed.

\begin{figure}[p] 
    \centering
    
    \includegraphics[width=0.31\textwidth]{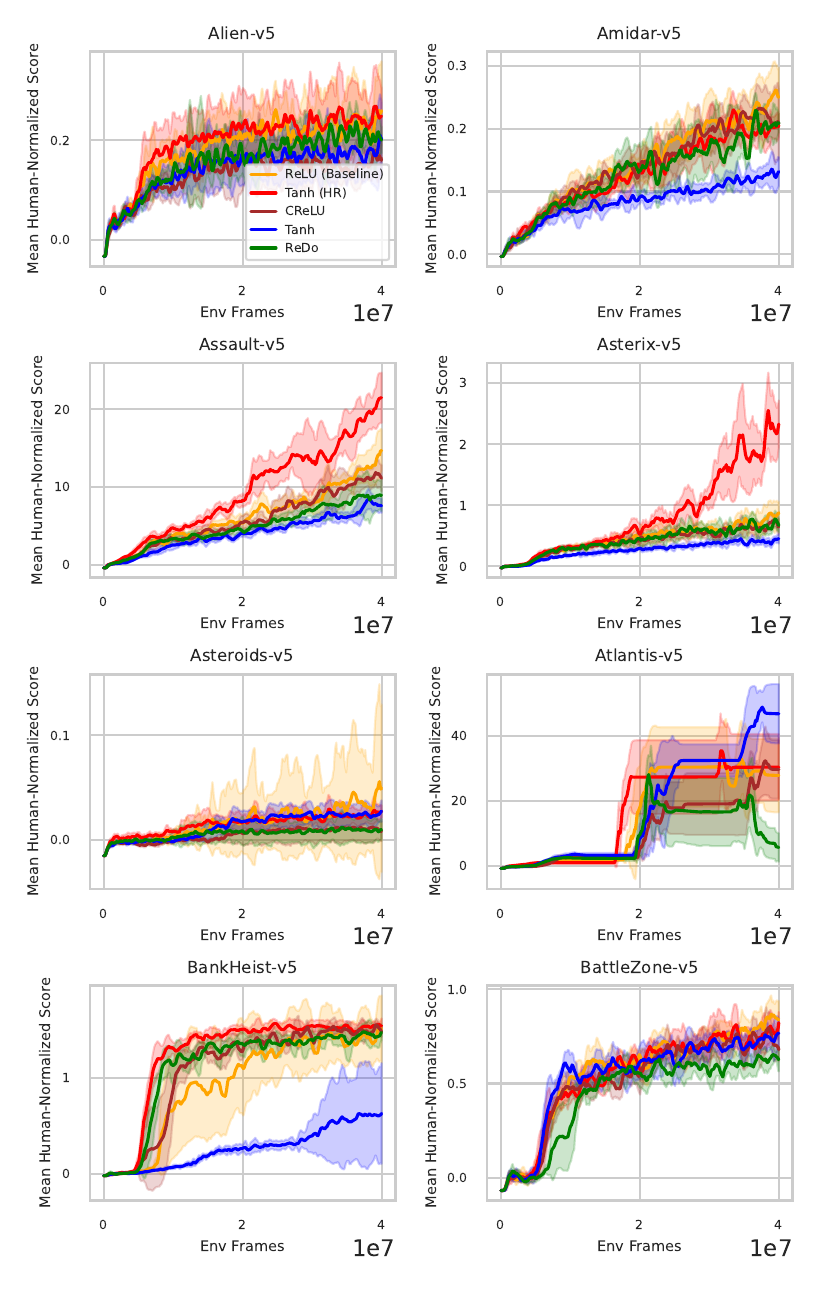}\hfill
    \includegraphics[width=0.31\textwidth]{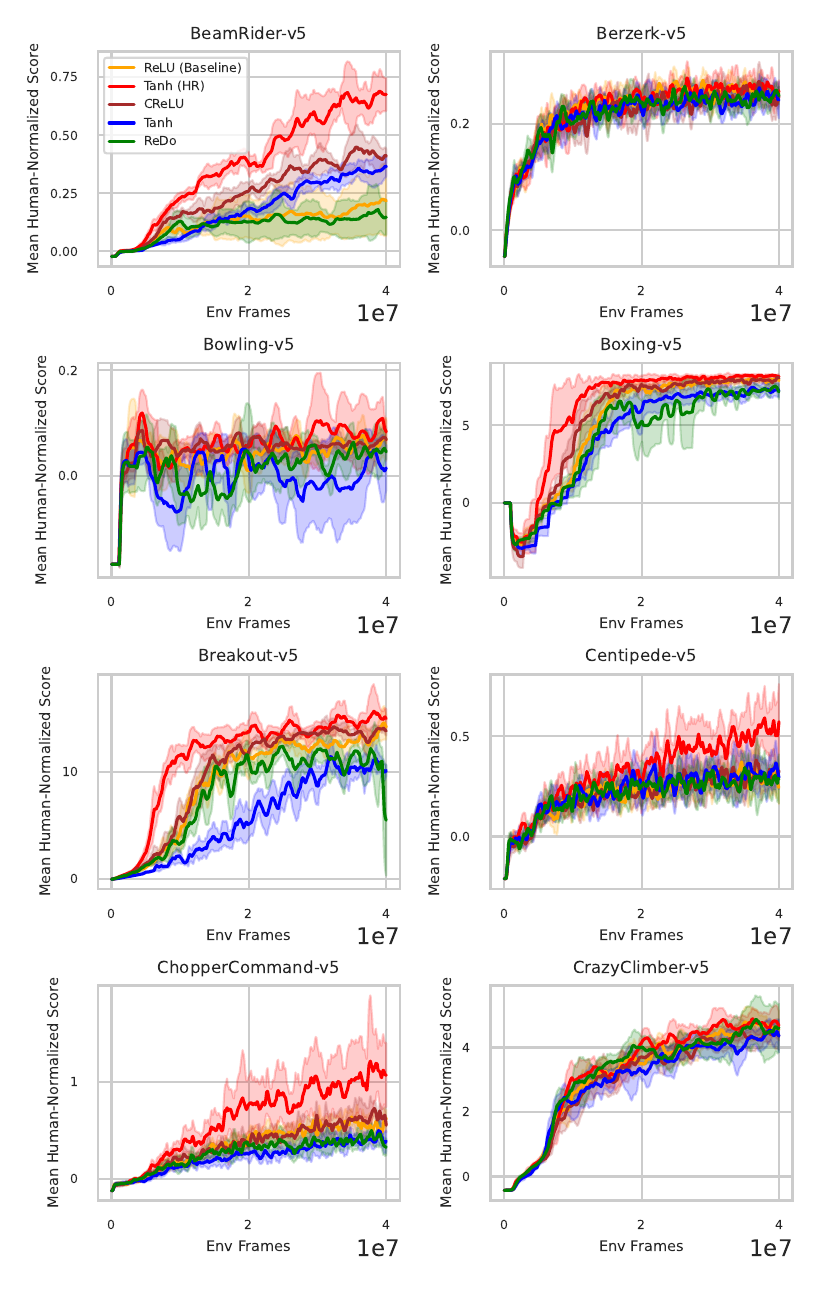}\hfill
    \includegraphics[width=0.31\textwidth]{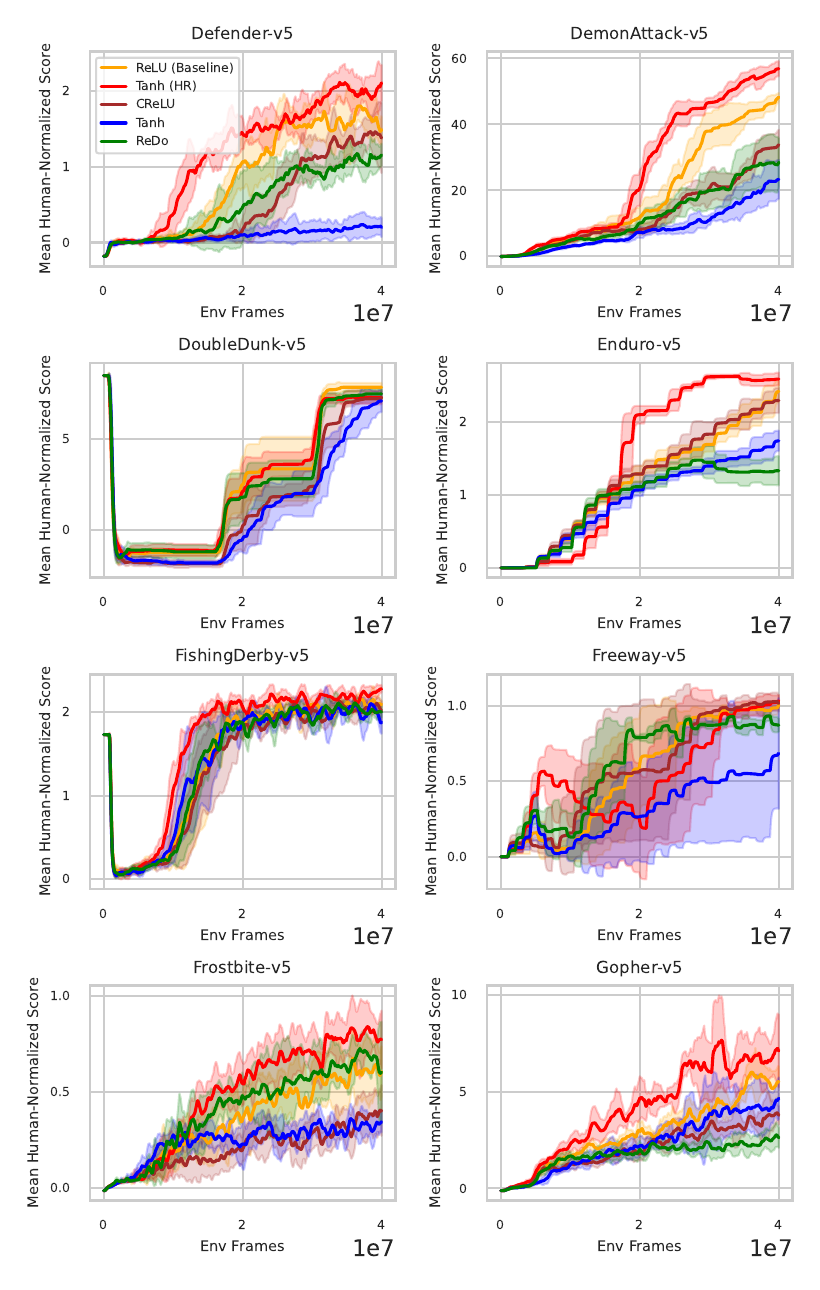}

    \vspace{0.5cm} 

    \includegraphics[width=0.31\textwidth]{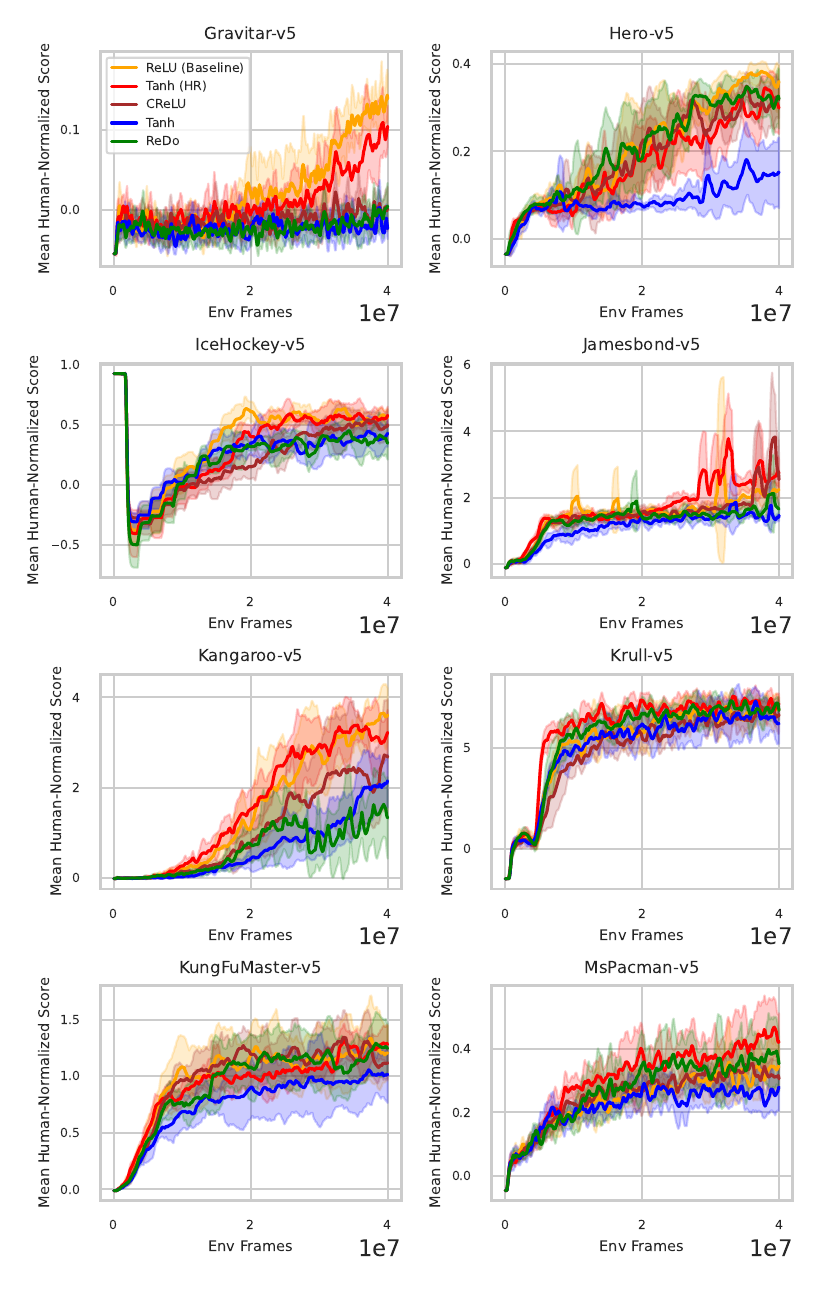}\hfill
    \includegraphics[width=0.31\textwidth]{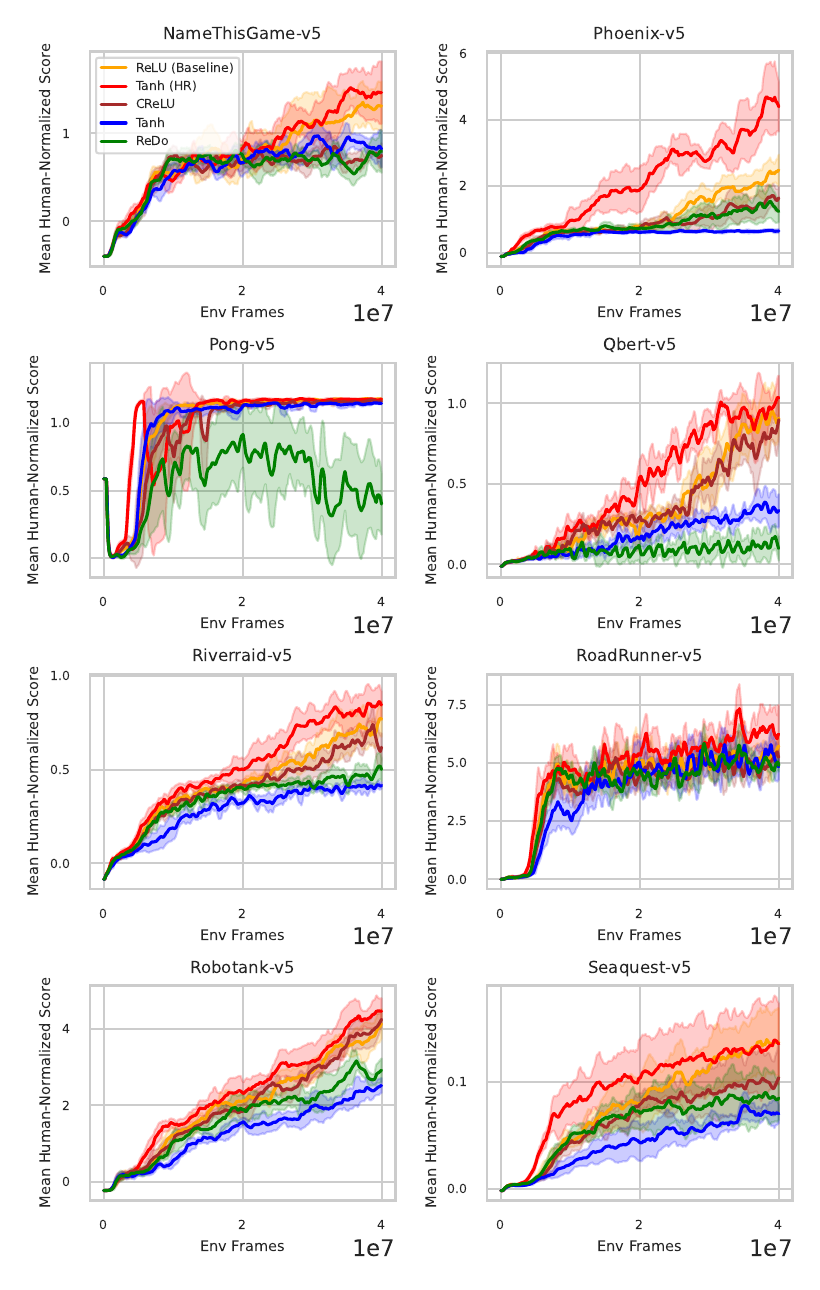}\hfill
    \includegraphics[width=0.31\textwidth]{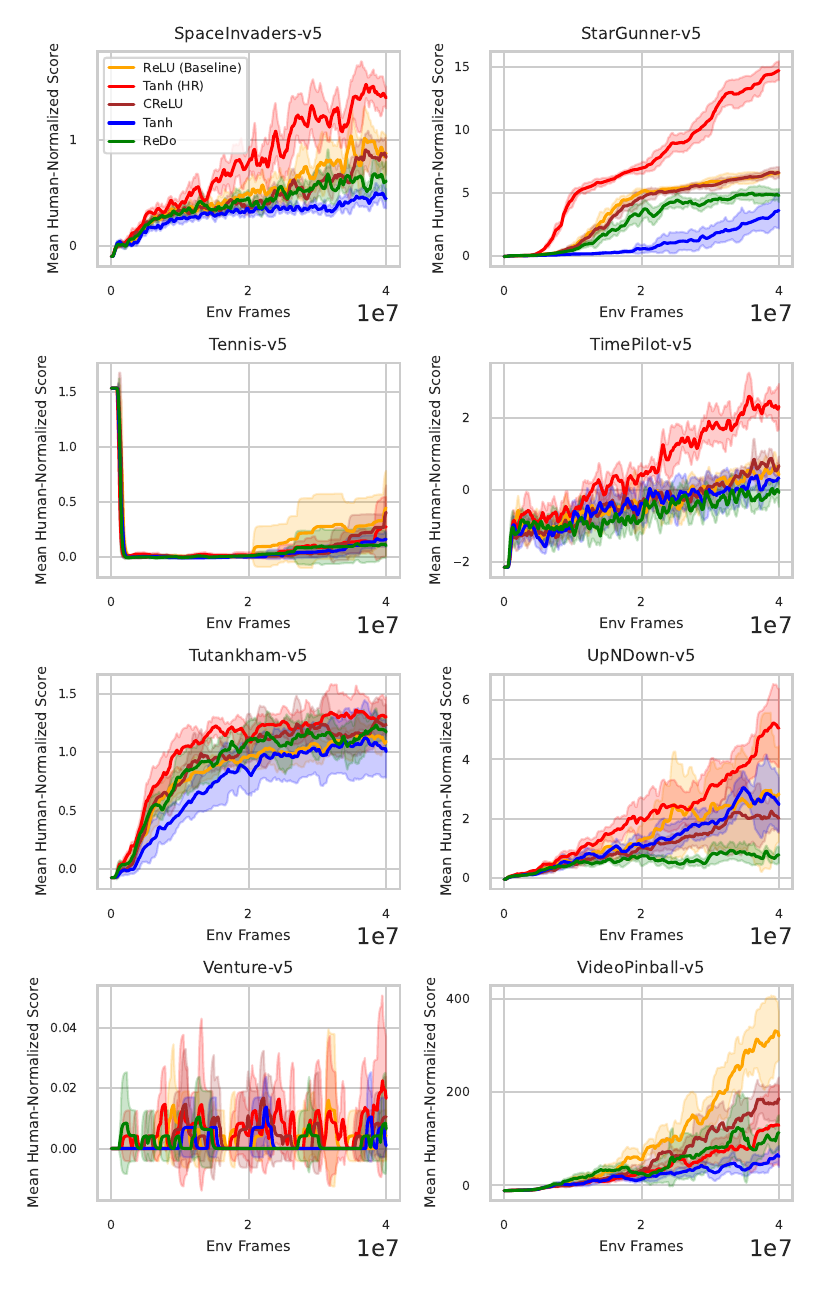}

    \vspace{0.5cm}

    \includegraphics[width=0.31\textwidth]{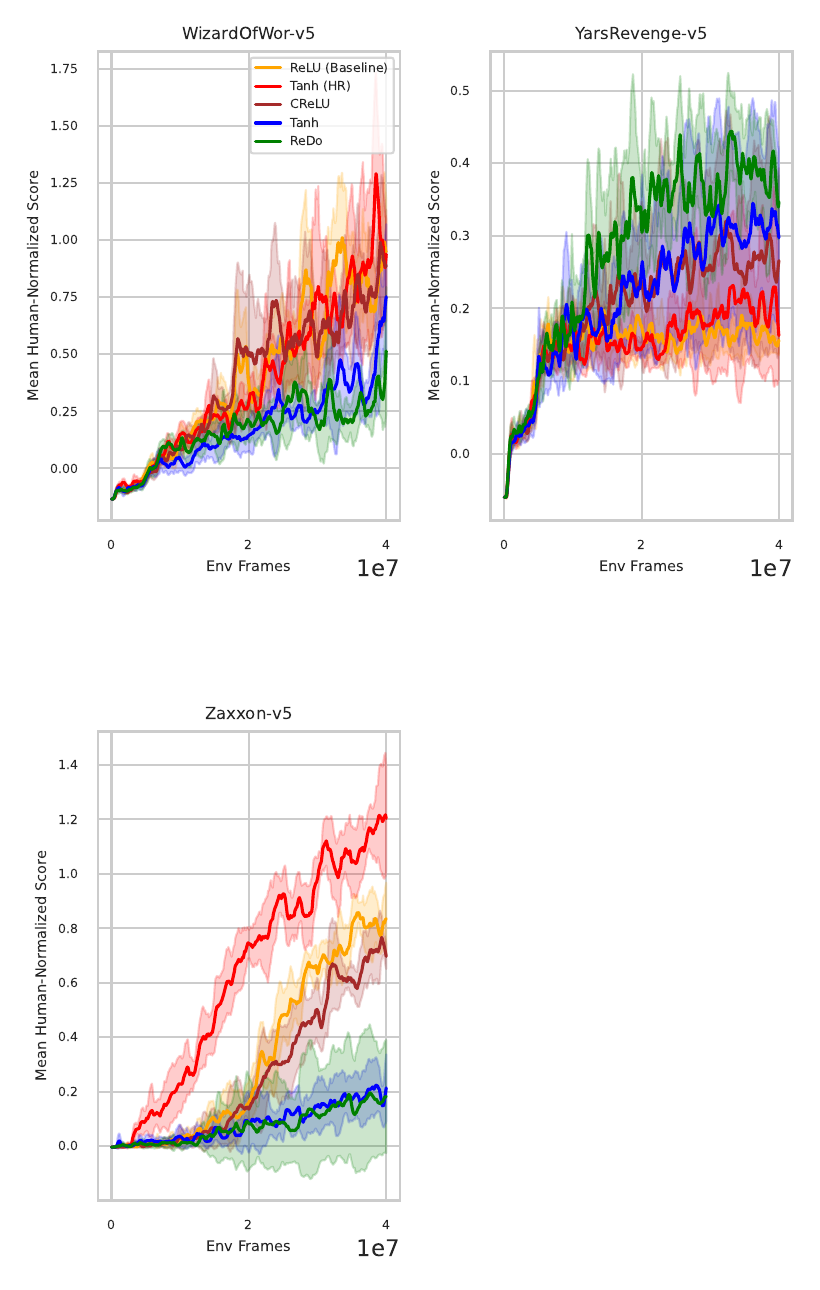}\hfill
    \begin{minipage}[b]{0.64\textwidth}
        \caption{PQN per-environment performance across 51 Atari games. Learning curves are averaged over 5 seeds; shaded regions show standard deviation.}
        \label{fig:pqn_individual_all}
    \end{minipage}

\end{figure}

\clearpage

\subsection{SimbaV2 on 14 HumanoidBench Tasks}
\label{app:hb}
For each HumanoidBench task $t$, we report a success-normalized score that linearly rescales the raw episodic return so that a random policy receives $0$ and the task's success threshold receives $1$:

\begin{equation}
    \tilde{r}_t = \frac{r_t - r^{\text{random}}_t}{r^{\text{target}}_t - r^{\text{random}}_t}
    \label{eq:hb-success-norm}
\end{equation}

where $r_t$ is the agent's mean evaluation return on task $t$ over the final 5\% of training, $r^{\text{random}}_t$ is the average return of a uniform random policy, and $r^{\text{target}}_t$ is the per-task success threshold defined by HumanoidBench~\citep{sferrazza2024humanoidbench}. 

An episode is considered successful by the benchmark whenever its return exceeds $r^{\text{target}}_t$. Thus, $\tilde{r}_t = 1$ corresponds to consistently solving the task, while $\tilde{r}_t > 1$ indicates super-target performance. The per-task values of $r^{\text{random}}_t$ and $r^{\text{target}}_t$ are listed in Table~\ref{tab:hb-norm-scores}.

To summarize across the 14 HumanoidBench tasks, we report the interquartile mean (IQM) of $\{\tilde{r}_t\}$, i.e., the mean after discarding the bottom and top 25\% of (task, seed) scores. Per-task curves are plotted as the mean over seeds with shaded standard error.

\begin{table}[h]
  \centering
  \caption{Per-task random and success-target scores for HumanoidBench used in Eq.~\eqref{eq:hb-success-norm}.}
  \label{tab:hb-norm-scores}
  \begin{tabular}{lrr}
    \toprule
    Task & Random & Target \\
    \midrule
    \texttt{h1-balance-simple} & 9.39   & 800   \\
    \texttt{h1-balance-hard}   & 9.04   & 800   \\
    \texttt{h1-crawl}          & 272.66 & 700   \\
    \texttt{h1-hurdle}         & 2.21   & 700   \\
    \texttt{h1-maze}           & 106.44 & 1200  \\
    \texttt{h1-pole}           & 20.09  & 700   \\
    \texttt{h1-reach}          & 260.30 & 12000 \\
    \texttt{h1-run}            & 2.02   & 700   \\
    \texttt{h1-sit-simple}     & 9.39   & 750   \\
    \texttt{h1-sit-hard}       & 2.45   & 750   \\
    \texttt{h1-slide}          & 3.19   & 700   \\
    \texttt{h1-stair}          & 3.11   & 700   \\
    \texttt{h1-stand}          & 10.55  & 800   \\
    \texttt{h1-walk}           & 2.38   & 700   \\
    \bottomrule
  \end{tabular}
\end{table}

\begin{figure}[!htb]
    \centering
    \includegraphics[width=0.9\linewidth]{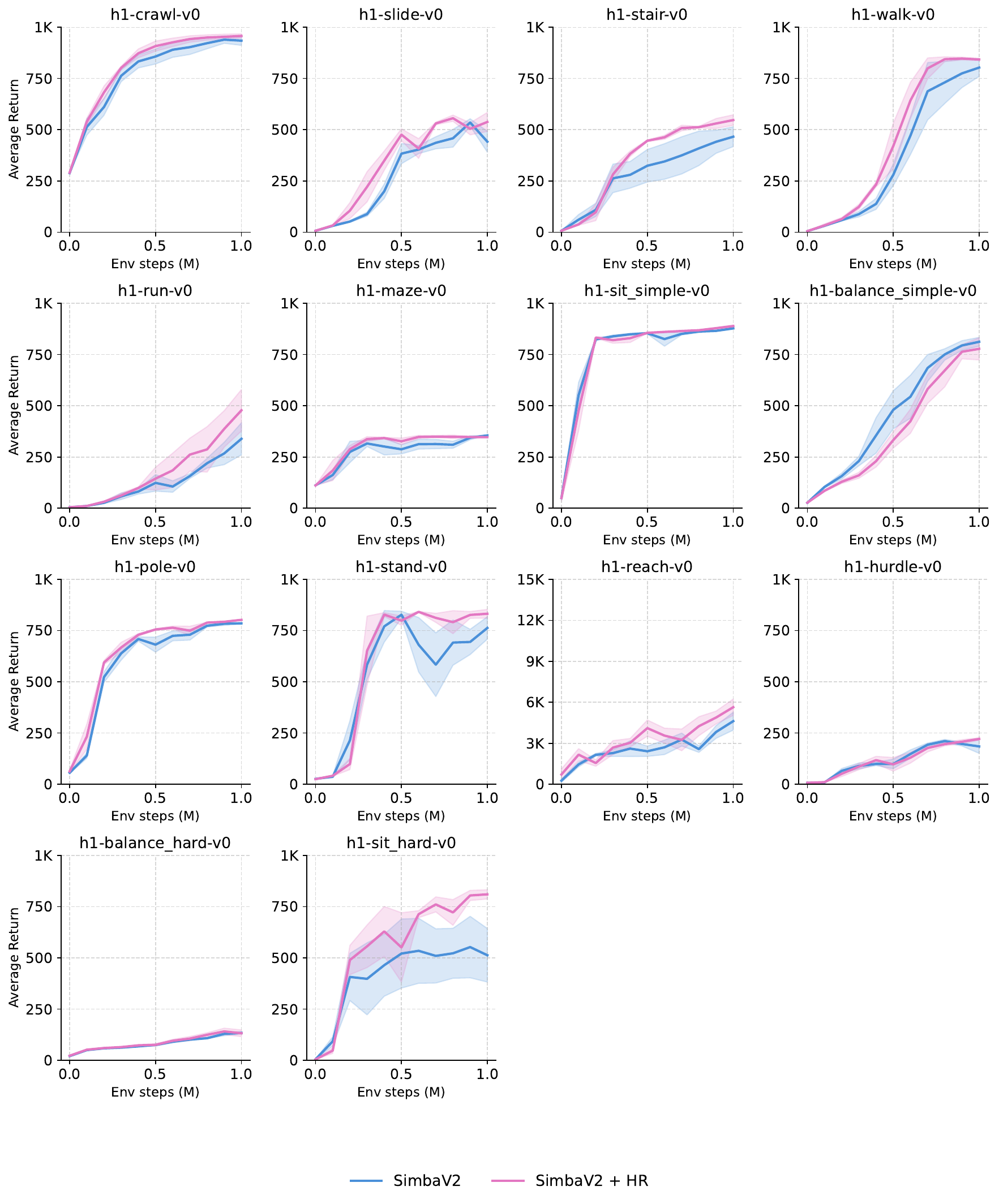}
    \caption{Performance on 14 HumanoidBench tasks. Learning curves are averaged over 3 seeds; shaded regions show standard error.}
    \label{fig:simbav2_all}
\end{figure}

\clearpage

\subsection{MR.Q on 28 DMC-Visual Tasks}
\label{app:dmc_visual}
\begin{figure}[!hbt]
    \centering
    \includegraphics[width=0.9\linewidth]{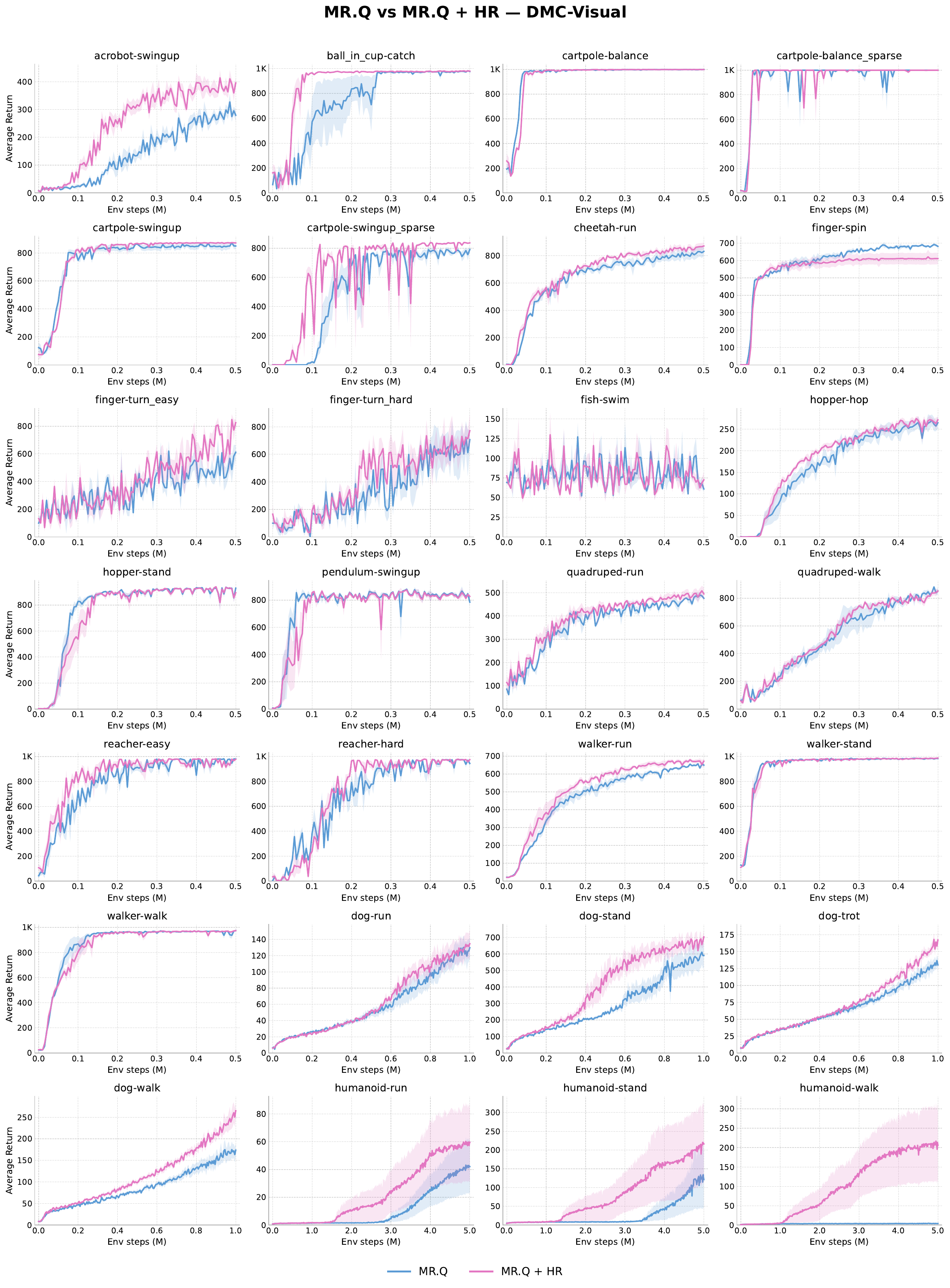}
    \caption{Performance on 28 DMC-Visual tasks. Learning curves are averaged over 3--5 seeds; shaded regions show standard error.}
    \label{fig:all_dmc_visual}
\end{figure}

DMC-Visual includes 28 tasks in total, 7 of them is categorized as \textbf{Hard}, i.e. 4 dog tasks and 3 humanoid tasks (see the last 2 rows in~\cref{fig:all_dmc_visual}).

\clearpage
\newpage

\end{document}